\documentclass[10pt,twocolumn,letterpaper]{article}

\usepackage{cvpr}              

\usepackage{algorithm}
\usepackage[noend]{algpseudocode}
\usepackage[accsupp]{axessibility}  

\definecolor{cvprblue}{rgb}{0.21,0.49,0.74}
\usepackage[pagebackref,breaklinks,colorlinks,allcolors=cvprblue]{hyperref}

\def\paperID{9} 
\def\confName{CVPR}
\def\confYear{2026}

\title{SeasonScapes: Learning Large-scale Re-lightable 3D Landscapes with Seasonal Variation from Sparse Webcams}

\author{Timo Kleger$^{2}$ \\
\and
Qi Ma$^{1, 2}$ \\
\and 
Deheng Zhang$^{1}$ \\
\and
Luc Van Gool$^{1}$ \\
\and
Danda Pani Paudel$^{1}$ \\
\and
$^{1}$INSAIT, Sofia University “St. Kliment Ohridski”
\and
$^{2}$ETH Zurich
}

\newcommand{\thickhline}{\specialrule{0.1em}{0.05em}{0.05em}}  

\begin{document}

\twocolumn[{%
    \maketitle
    \vspace{-1em}
    \begin{center}
        \includegraphics[width=\linewidth]{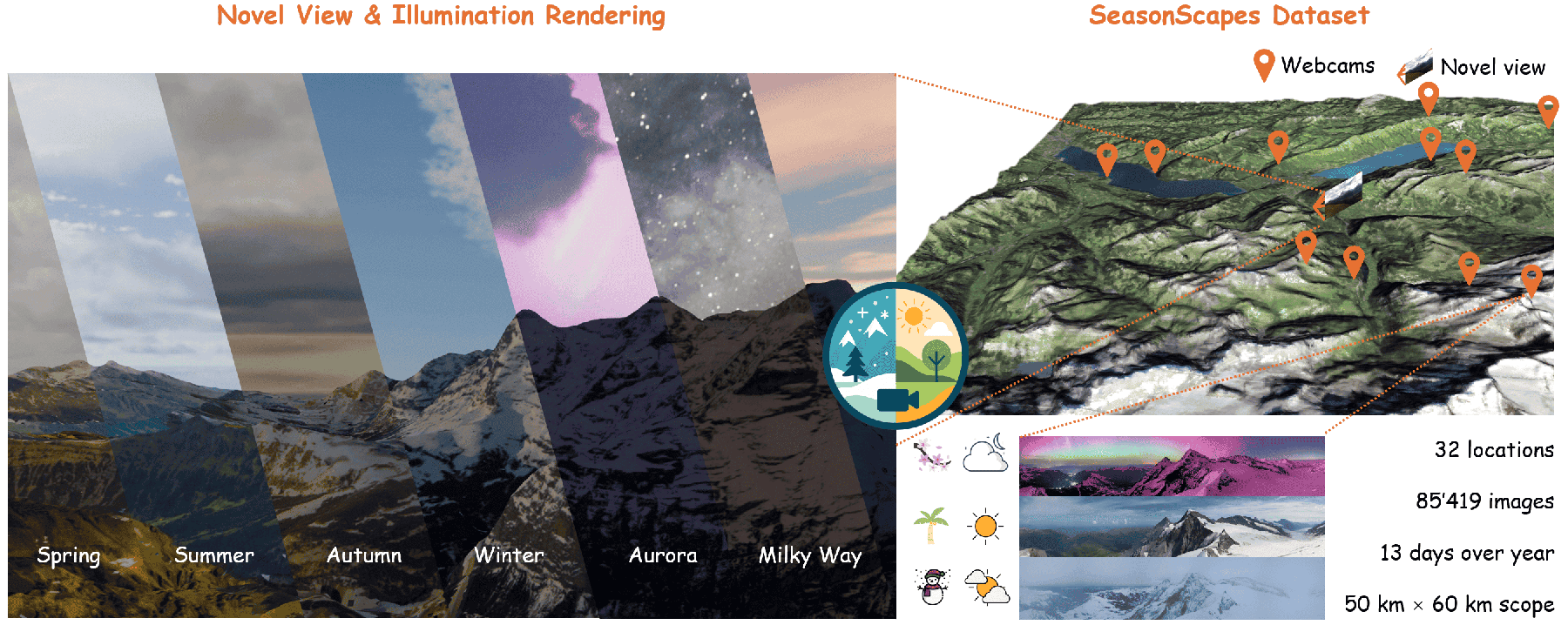}
        \captionof{figure}{We introduce SeasonScapes framework and a the SeasonScapes dataset: Swiss Sparse-view Mountain Scenes with Seasonal Changes that covers over 50 km $\times$ 60 km, composed of more than 85,000 webcam images captured from 32 different locations across 13 timestamps throughout a full year. By projecting these timestamp-specific images onto a 3D mesh, we construct seasonal 3D landscapes that reflect natural appearance changes over time. To address occlusions and missing data, we leverage conditional diffusion models for image-guided inpainting directly on the mesh. The resulting completed meshes can be further relighted using standard physically-based renderer.}
        
        \label{fig:teaser}
    \end{center}
}]

\begin{abstract}
  Large-scale scene creation from sparse views in the wild remains an open and significant problem in computer vision due to challenges such as camera calibration errors, limited scene coverage, and high scene complexity, including lighting and appearance variations. To address this, we introduce a digital twin of the Swiss alpine landscape, capturing temporal variations across seasons and times of day using our newly proposed SeasonScapes dataset for sparse-view mountain scenes with seasonal changes. We associate sparsely distributed webcams across 3000\,$\text{km}^2$ of the Bernese Highlands with a low-resolution satellite image and elevation map via 2D–3D correspondence optimization. To overcome limited scene coverage, we propose a 3D mesh painting method that inpaints novel views using priors from the ControlNet pipeline. Unlike prior works that focus on centered, bounded objects, our approach handles large-scale, unbounded scenes by introducing linear or cubic spline inpainting trajectories between GPS coordinates and performing occlusion-aware painting via UV indexing through depth rendering – rather than relying solely on normal-based filtering. By running the pipeline across varying times of day and seasons, we create a dataset with rich appearance and lighting diversity. Using this data, we train a relightable Gaussian Splat model with appearance Multi-Layer Perceptron.
\end{abstract}
    
\section{Introduction}
\label{sec:intro}


Large-scale 3D landscape reconstruction holds substantial importance for real-world applications such as environmental monitoring, weather modeling, virtual tourism, infrastructure planning and disaster response. However, this is a challenging task in computer vision and graphics due to the high variance of the weather, season, and illumination. Beyond traditional mesh-based systems like Google Earth that rely on static geometry and blended satellite texture maps, our work focuses on achieving photometrically consistent 3D representations capable of modeling diverse illumination conditions and view-dependent appearance variations. We leverage known 3D geometry and sparse webcam imagery to reconstruct up-to-date scene states with high visual fidelity.

Existing large-scale 3D datasets primarily target on static urban environments~\cite{li2023matrixcity,Turki_2022_CVPR,Geiger2013IJRR}, with limited coverage of natural landscapes. These dataset have siginificantly advanced task like large-scale reconstruction in terms of efficiency, hierarchical scene representation~\cite{Turki_2022_CVPR,liu2025citygaussian,hierarchicalgaussians24} and task like scene understanding and localization \cite{ma2025cityloc,feng2025citygptempoweringurbanspatial}. To address this gap, we introduce the SeasonScapes dataset for sparse-view mountain scenes with seasonal and intraday illumination change, covering 3,000 \,$\text{km}^2$ of alpine Swiss terrain across 12 months to capture seasonal variations and complete diurnal cycles from dawn to dusk. This spatially and temporally comprehensive dataset enables modeling of both long-term environmental transitions and short-term illumination changes in natural mountain landscapes. We expand upon NeRF-W's temporal modeling dataset~\cite{martinbrualla2020nerfw} in scope and capability by offering: (1) large-scale natural landscapes rather than isolated urban landmarks; (2) precise timestamp annotations for all webcam observations; (3) calibrated 2D-3D correspondences between ground-level imagery and georeferenced satellite data. 

Although prior works~\cite{kulhanek2024wildgaussians3dgaussiansplatting, nerf_on_the_go, martinbrualla2020nerfw} tackle 3D reconstruction in unconstrained environments, they do not target large-scale landscape reconstruction from sparsely distributed camera views. We introduce SeasonScapes, a novel framework for reconstructing large-scale, relightable 3D landscapes that capture both seasonal and diurnal variations. Building upon our multimodal dataset, SeasonScapes integrates: macro-scale environmental structures (terrain topography, hydrographic boundaries, and urban infrastructure) with micro-scale dynamic observations (temporal illumination changes, biological activity patterns, and subsurface features). This approach advances beyond satellite-only reconstruction methods~\cite{automatic_3DReconstruction_satellite, satellite_3DReconstruction}, enabling comprehensive spatiotemporal scene understanding across multiple scales.

Our framework generates temporally consistent 3D landscapes with photorealistic lighting—spanning day/night cycles and seasonal variations from summer to winter—using arbitrary unconstrained webcam inputs. The pipeline begins by texturing the base mesh through projective blending of available 2D observations, then completes missing regions with depth-conditioned diffusion models that leverage optimized inpainting trajectories to preserve spatial consistency. Finally, we train a relightable Gaussian model that employs a multi-layer perceptron to capture dynamic scene properties. Through comprehensive experiments across three diverse test scenes, we demonstrate our method’s ability to render realistic environments while accurately modeling scene dynamics across varying lighting and seasonal conditions.

Our contributions can be summarized as: 

$\bullet$ We present SeasonScapes dataset, a high-quality large-scale multimodal dynamic dataset with seasonal and diurnal variations, covering 3,000 \,$\text{km}^2$ of alpine terrain.

$\bullet$ We propose SeasonScapes framework for relightable 3D landscape creation that leverages diffusion priors to inpaint unseen regions with high fidelity.

$\bullet$ We incorporate dynamic embeddings into relightable gaussian training, enabling effective modeling of temporal changes in large-scale outdoor scenes.

The source code, configurations, rendered examples and where to download the SeasonScapes dataset are available at: https://github.com/ChlaegerIO/SeasonScapes.

\section{Related Work}



\noindent{\textbf{Large-scale Landscape dataset.}}
Early work \cite{namin2015multi,Geiger2013IJRR, geo_localization_alps,geo_pose_cityScape} introduced paired 2D-3D semantic data for scene understanding, but primarily focused on urban and vehicular environments. The RUGD dataset \cite{RUGD2019IROS} centers on off-road semantic segmentation, subsequent works \cite{jiang2020rellis3d,wildscenes2024} extend this to 3D scene understanding. However, these datasets remain limited in scale, covering less than 20 km of trajectory data. \cite{brejcha2020landscapear} also introduced dataset featuring alpine region elevation models, demonstrating robust cross-domain camera pose estimation through learned descriptor matching. \cite{wildscenes2024} benchmark the 2D and 3D semantic segmentation. Most closely related to our work is \cite{brejcha2020landscapear} which introduced a dataset featuring alpine region elevation models and demonstrated robust cross-domain camera pose estimation through learned descriptor matching.


\noindent{\textbf{Sparse-view mesh colorization.}} 
Traditional methods for synthesizing textures on 3D assets typically relied on placing hand-crafted patterns or performing global optimization \cite{texture_traditional1, texture_traditional2, texture_traditional3}. More recently, learning-based approaches have demonstrated strong performance in generating plausible textures for complex 3D shapes \cite{texture_learning1, texture_learning2, texture_learning3, texture_learning4}. One notable method, TEXTure \cite{texture_3d}, iteratively applies a pretrained depth-to-image diffusion model to paint the texture map from multiple viewpoints, conditioning each step on previous results. While effective, this approach lacks global context and often produces view-inconsistent textures. TexFusion \cite{textFusion} addresses this limitation by aggregating texture information across multiple views during the denoising process, thereby improving view consistency. Text2Tex \cite{text2text} further automates viewpoint selection, reducing manual intervention. Despite these advancements, existing methods still suffer from lighting bias introduced by 2D priors, leading to inconsistent textures under varying illumination. To mitigate this, Paint3D \cite{zeng2023paint3dpaint3dlightingless} introduces a refinement model trained on illumination-free texture data, significantly reducing lighting artifacts and improving overall consistency. We adapt these diffusion-based texture inpainting techniques to the challenging setting of large-scale landscape reconstruction, achieving photorealistic and view-consistent rendering across varying illumination and viewpoints.

\noindent{\textbf{3D Scene editing and relighting.}} Recent advancements in novel view synthesis \cite{nerf_novel_view, novel_view_mip_nerf, 3DGS, novel_view_mipSplatting} have primarily concentrated on static environment reconstruction. Subsequent works \cite{lazova2023control, kania2022conerf, wang2022clip, zhang2024coarfcontrollable3dartistic, yu2023pointbased, zheng2024gaussiangrasper} have extended this toward 3D scene editing and controllability. Some approaches enable object-level relighting by leveraging 3D distillation of diffusion model \cite{zhao2024illuminerf, jin2024neural_gaffer} or inverse rendering \cite{munkberg2022extracting, boss2021nerd, jin2023tensoir, liang2024gsir, zhang2025rise}, while others address scene-level illumination editing \cite{zhu2023i2, wu2025gsssr, liang2024gusirgaussiansplattingunified}. However, current scene-level relighting methods often struggle to reconstruct the scene under diverse lighting conditions. WildGS \cite{kulhanek2024wildgaussians3dgaussiansplatting} addresses this by incorporating DINO v2 features \cite{oquab2024dinov2learningrobustvisual} and per-image appearance embeddings \cite{martinbrualla2020nerfw}, enabling faster optimization and robust handling of dynamic scenarios, including illumination changes. These per-image features can further be leveraged to edit lighting conditions at render time. Despite these advances, methods based on 3D Gaussian Splatting (3DGS) \cite{3DGS} often suffer from view overfitting and degrade in quality under sparse camera views, such as those captured from webcams. To address this, our proposed pipeline introduces camera view augmentation.


\begin{figure*}[!thb]
    \centering
    \includegraphics[width=1\linewidth]{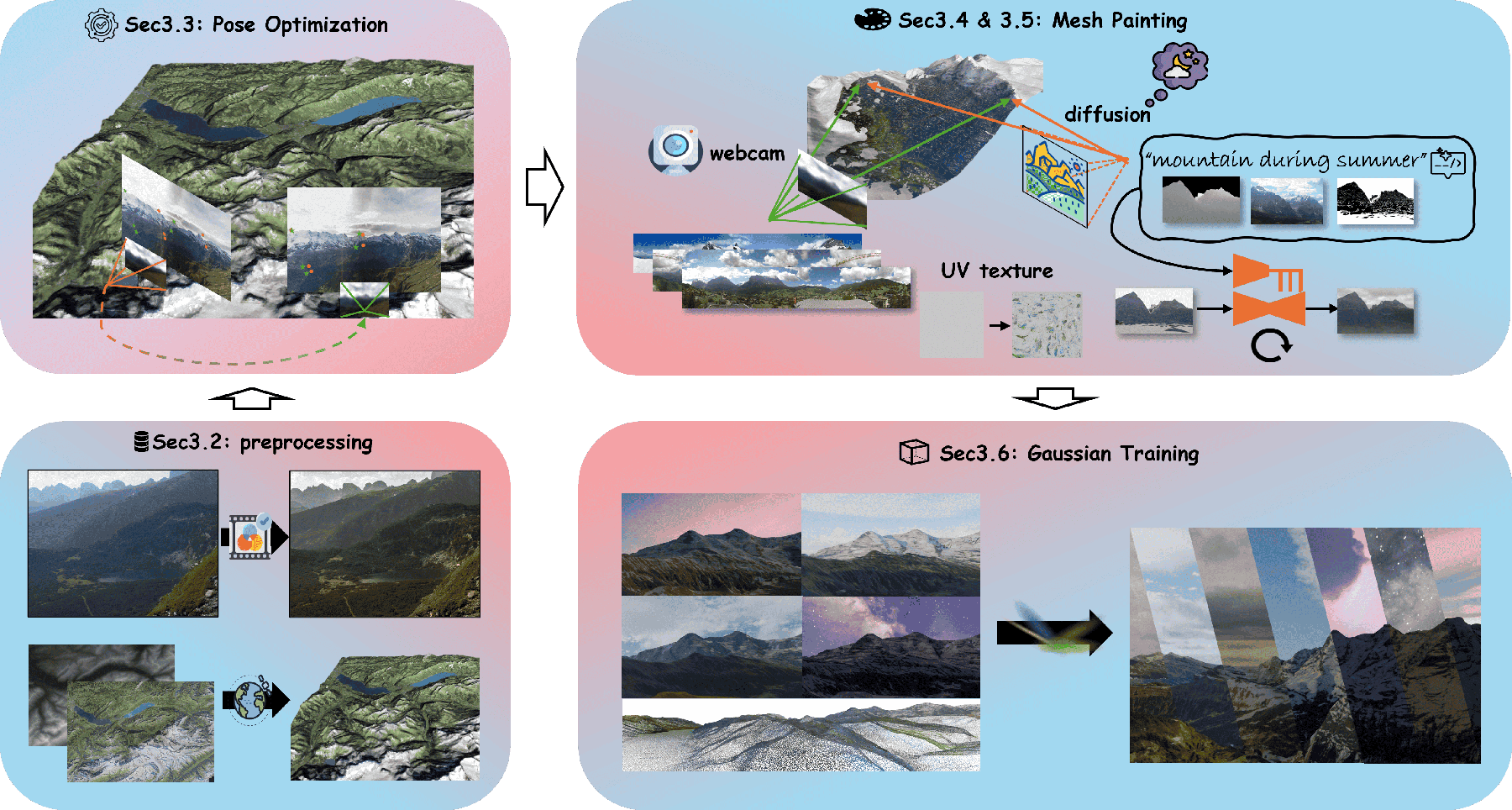}
    \caption{\textbf{Pipeline overview.} Our pipeline generates time-varying 3D landscapes through a multi-stage process. First, we preprocess Google Earth data and webcam imagery (Sec. \ref{sec:preprocessing}). We then employ a learning-based approach to optimize camera parameters, aligning the 2D webcam frames with the 3D landscape point cloud (Sec. \ref{sec:camOptim}). In the landscape painting stage, we project the preprocessed images to texture the UV map (Sec. \ref{sec:painting}). To ensure completeness, we perform mesh inpainting (Sec. \ref{sec:inpainting}), using an off-the-shelf ControlNet to iteratively fill occluded or missing regions. Finally, after post-processing the UV maps, the outputs from multiple runs are used to supervise the seasonal and diurnal embeddings of a WildGaussian model.}
    \label{fig:framework}
\end{figure*}

\section{Method}
\label{sec:method}
\subsection{Preliminary}

\noindent{\textbf{Diffusion Models.}}
Diffusion models~\cite{ho2020denoising,sohl2015deep} learn complex data distributions by reversing a multi-step noising process. The noising process transforms data samples $x$ into noise over a sequence of $T \in \mathbb{N}$ steps. The model is trained to learn the denoising process.

A Denoising Diffusion Probabilistic Model (DDPM) applies Gaussian noising through $T$ steps with variance schedule $\beta_1, ..., \beta_T$ where:
\begin{equation}\label{eq:q}
q(x_t \mid x_{t-1}) = \mathcal{N}(x_t; \sqrt{1 - \beta_t} x_{t-1}, \beta_t \mathbb{I}),
\end{equation}
where $\mathbb{I}$ is the identity matrix.
This schedule ensures $x_T$ approaches an isotropic Gaussian, \ie, $q(x_T) \approx \mathcal{N}(\mathbf{0}, \mathbb{I})$.
Setting $\alpha_t = 1 - \beta_t$ and $\bar{\alpha}_t = \prod_{i=1}^t \alpha_i$ yields a closed-form solution for directly sampling $x_t$ given a data point $x_0$: 
\begin{equation}
x_t \sim q(x_t \mid x_0) = \mathcal{N}(x_t; \sqrt{\bar{\alpha}_t} x_0, (1 - \bar{\alpha}_t) \mathbb{I}).
\end{equation}

For sufficiently small $\beta_t$, the reverse $p_\theta(x_{t-1} \mid x_t)$ is Gaussian.  
Thus, we approximate it with model $\mathcal{F}_{\theta _d}$:
\begin{equation}\label{eq:p}
p_\theta(x_{t-1} \mid x_t) = \mathcal{N}(x_{t-1}; \sqrt{\alpha_t} \mathcal{F}_{\theta_d}(x_t, t), (1 - \alpha_t) \mathbb{I}).
\end{equation}

\noindent{\textbf{ControlNet.}} Recent advancements to steer diffusion models based on conditional inputs have yielded to astonishing results \cite{ControlNet, adaptors_ip, controlingDiffusion_mask, adaptors_T2I, controlingDiffusion_textPromptOneWord}. One of which is the ControlNet which adds an additional fine-tunable encoder block $F_{\theta_c}$ and zero convolution $\mathcal Z$ to the diffusion model $\mathcal{F}_{\theta_d}$ 
\begin{equation}
    \mathcal F_\theta = \mathcal F_{\theta_d}(x_t, t) + \mathcal Z(\mathcal F_{\theta_c}(x_t + \mathcal Z(c))).
\end{equation}
The additional encoder block is added to the standard diffusion model with a zero convolution to start fine-tuning from the diffusion model. The additional condition $c$ such as depth, normals, human pose or other control images helps to guide the diffusion model to the desired output.


\noindent{\textbf{3D Gaussian Splatting.}}
\label{subsec:gs_preliminary}
3DGS represents scene space with Gaussian primitives $\{Y_i\}_{i=1}^{N}$, stacking these as follows:
\begin{equation}
Y = [C, O, S, R, SH] \in \mathbb{R}^{N\times59},
\end{equation}
where $C \in \mathbb{R}^{N\times3}$ denotes the centroid, $O \in \mathbb{R}^{N\times1}$ the opacity, $S \in \mathbb{R}^{N\times3}$ the scale, $R \in \mathbb{R}^{N\times4}$ the quaternion rotation vector, and $SH \in \mathbb{R}^{N\times48}$ the degree-3 spherical harmonics. Each Gaussian defines a spatial area with opacity. Scene point $q$ is influenced by Gaussian $Y_i$ via opacity-weighted distribution where: 
\begin{equation}
h_{i}(q) = O_i \exp\left( -\frac{1}{2} (q - C_{i})^T \Sigma_{i}^{-1} (q - C{i}) \right),
\end{equation}
where covariance $\Sigma_{i}$ is formulated as $\Sigma_{i} = R_i S_i S_i^{T} R_i^{T}$.

Projected onto a 2D image plane, each Gaussian’s influence, $h$, contributes to a pixel’s color through an alpha-blending equation over the set $\mathcal{G}$ of influencing Gaussians:
\begin{equation}
c_\text{pixel} = \sum_{i \in \mathcal{G}} c_i h^{\textrm{2D}}_{i} \prod_{j=1}^{i-1} (1 - h^{\textrm{2D}}_{j})\enspace.
\end{equation}

Differentiable rasterization enables gradient-based optimization of Gaussian parameters. In this manner, we represent images rendered from pose $P$ given by $I=\mathcal{G}(P)$.


\subsection{Data Preprocessing}
\label{sec:preprocessing}

We begin by preprocessing the panoramic webcam images and the 3D elevation model as follows:

\noindent{\textbf{Initialize 3D Landscape.}} We start from a low-resolution Digital Elevation Model $I_{DEM}$ and an aerial satellite color image $I_S$ of our specified region that we downloaded using Google Earth Engine \cite{EarthEngineAPI}. Then we create a 3D point for each pixel using the normalized height from $I_{DEM}$ and color it with the satellite image $I_S$. Based on the point cloud we create a mesh using the Poisson surface reconstruction algorithm \cite{PoissonMesh} $M = (V, F)$ with vertices $V = \{v_i\}, v_i \in \mathbb R^3$ and triangular faces $F = \{f_i\}, f_i \in \{v_i\}^3$ in the operation $\mathcal M: (I_{DEM}, I_S) \mapsto M.$
    

We unwrap the surface of our 3D mesh using the library xatlas \cite{xatlas_library} and get the UV texture image $T \in \mathbb R^{H\times W \times 3}$ with a predefined resolution. Using a UV map allows texture and appearance information to be stored separately from geometry, enabling efficient reuse across large and complex scenes. 

\noindent{\textbf{Webcam Preprocessing.}} Most selected panorama webcams employ a 360° horizontal field of view with cylindrical projection. To align with ControlNet’s planar image processing, we split each webcam feed into 4–6 perspective sub-images facing distinct directions, followed by projection to planar coordinates. For better visualization we additionally color graded the images in Lightroom by adding more contrast, reducing the highlights, elevated shadow levels, and modified white balance toward warmer tones.



\subsection{Camera Parameter Optimization}
\label{sec:camOptim}
\begin{figure}[!thb]
    \centering
    \includegraphics[width=1\linewidth]{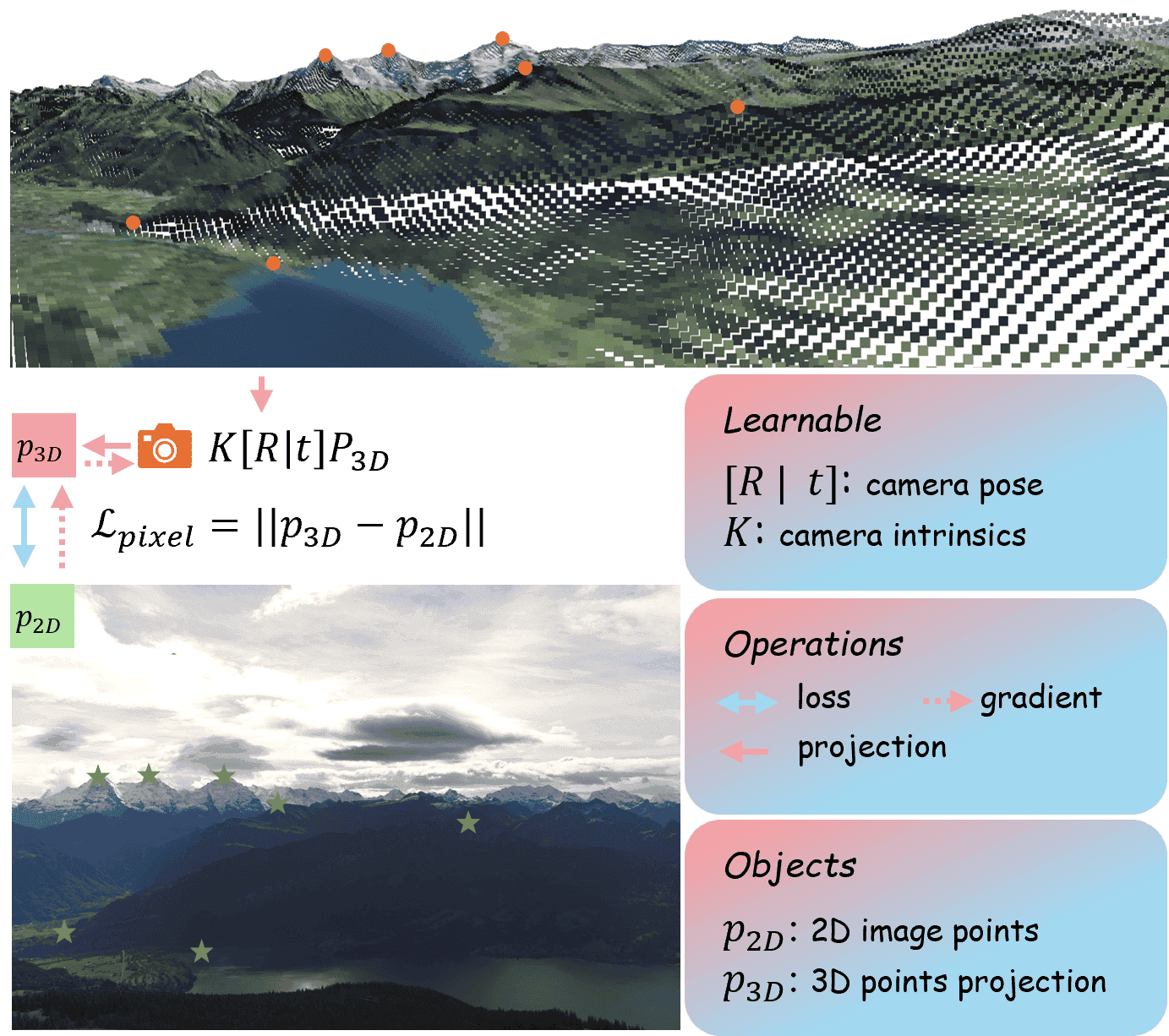}
    \caption{\textbf{Camera parameter optimization.} We refine the initial camera intrinsics and extrinsics (derived from GPS coordinates) by calculating the $\mathcal L_1$ pixel loss between the 3D points projection (orange circle) with manually labeled correspondent 2D image points (green). }
    \label{fig:camera_poseOptim}
\end{figure}

As shown in \cref{fig:camera_poseOptim}, given the preprocessed 3D mesh, we associate 3D points with 2D webcam images. While initial poses are estimated from RoundShot camera GPS coordinates, calibration errors persist due to perspective differences (parallax in terrain), limited GPS accuracy ($\pm $3–10m), elevation datum mismatches and temporal drift between satellite/webcam captures. To address this, we first manually establish 3D-to-2D correspondences using custom tools developed with Tkinter. For each matched pair $(p_\text{3D}, p_\text{2D})$,  we optimize learnable camera parameters by minimizing the $\mathcal L_1$ loss between projected 3D points $p_\text{3D}$ and target 2D points:
\begin{equation}
    \mathcal L_{\text{pixel}} = ||p_{3D} - p_{2D}||_1.
\end{equation} 
Since the webcam images are already undistorted, we model the projection using a pinhole camera projection as follows:
\begin{equation}
    p_\text{3D} = \Pi(P_\text{3D}) = \begin{bmatrix}
        u \\
        v\\
        1
    \end{bmatrix} = \begin{bmatrix}
        f_x  & 0  & c_x \\
        0    & f_y & c_y \\
        0    &  0  & 1
    \end{bmatrix} \begin{bmatrix}
        R  & t \\
        0  & 1
    \end{bmatrix} P_{3D}
\end{equation}
with focal lengths $f_x$ and $f_y$, principle point $(c_x, c_y)$, rotation $R \in \mathbb R^{3\times 3}$ and translation $t\in \mathbb R^3$. We optimize the intrinsic camera parameter $f_x$ and $f_y$ and the extrinsic camera parameter for the horizontal rotation. 
%

\subsection{Webcam Texturing in UV Space}
\label{sec:painting}
Our proposed 3D SeasonScapes starts with painting the mesh with our preprocessed webcams $I_w$ (section~\ref{sec:preprocessing}) at the optimized pose $p_w$ (section~\ref{sec:camOptim}) and the initialized 3D mesh $M_0$ with the corresponding texture map $T_0$.

\noindent{\textbf{Initial Viewpoint.}} We texture the UV map $T_1$ (Alg. 1 line 4, 1.4) with the first webcam image $I_1$ using the pose $p_1$ and the seen face indices $f_1^{(idx)}$ (Alg. 1.3), denoted as 
\begin{equation}
    \mathcal T_1: (M_0, I_1, p_1, f_1^{(idx)}, m_0^{UV}) \mapsto T_{1}.
\end{equation}
The UV mask $m_0^{UV}$ is one everywhere for the first image.

\noindent{\textbf{Painting Viewpoints.}} The next webcam poses $p_t$ are executed in a similar process $\mathcal T_t$ (Alg. 1.4) while taking into account the already textured regions $T_{[1, t-1]}$. Specifically we calculate a UV texture mask $m_{t-1}^{UV}$ of regions in the UV map that are already painted. And then we superpose all previous textures $T_{[1,t-1]}$ with the current texture $T_c$ 
\begin{equation}
    T_t = m_{t-1}^{UV} \odot T_{t-1} + (1-m_{t-1}^{UV}) \odot T_c.
    \label{eq:texturing}
\end{equation}
Therefore the painting is progressively added to the previous texture following Fig.~\ref{fig:pipeline_inpainting} without the inpainting ControlNet and it's inputs until all $W$ webcam views are painted $T_W = T_{[1,W]}$.

\begin{figure}[!thb]
    \centering
    \includegraphics[width=1.0\linewidth]{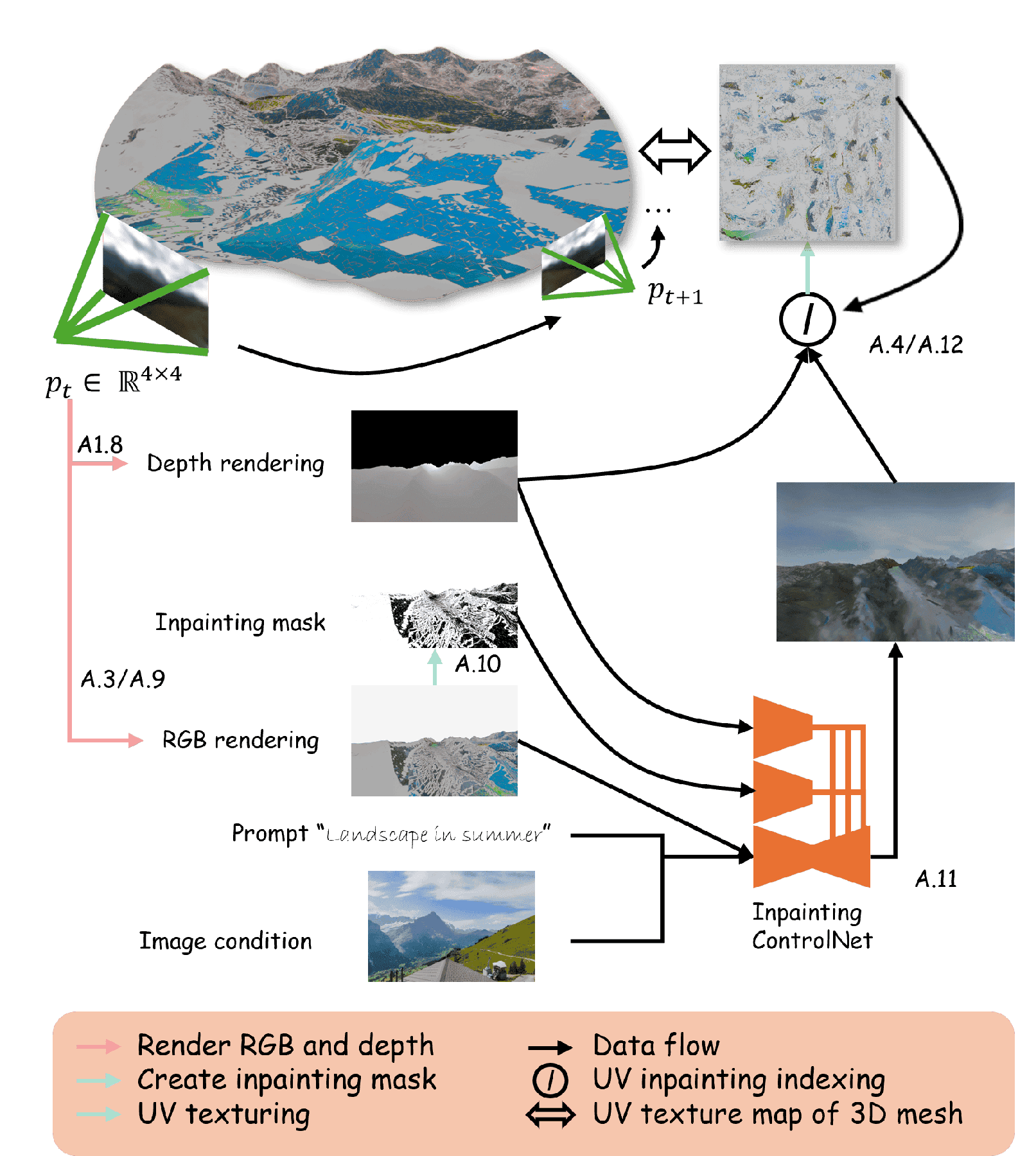}
    \caption{\textbf{3D mesh texturing including inpainting.} We iterate over the entire inpainting trajectory $\{p_t\}_{t=1}^{N}$. For each step $t$ we render an RGBD image. Using the RBG image we mask unseen regions using the HSV color space. We condition ControlNet using the inpainting mask, depth map, text prompt, and IP-Adapter image. The diffusion output is projected onto the UV map via depth-indexing of visible regions. The algorithm description A1.8 refers to the 8 line of the algorithm~\ref{alg:seasonMesh}.}
    \label{fig:pipeline_inpainting}
\end{figure}

\subsection{Iterative Inpainting}
\label{sec:inpainting}
In the second step we inpaint missing regions using a depth-aware 2D inpainting diffusion model along a user-specified novel view path with poses $\{p_t\}_{t=1}^N$. We start from the webcam painted texture $T_0 = T_W$ i.e. mesh $M_W$ (Alg. 1.5). 
As in the painting step we render the depth $d_t$ and visible faces $f_t^{(idx)}$. Additionally we render a partially colored RGB image $\Tilde{x}_t$. The rendering process is described as $\mathcal R: (M_{t-1}, p_t) \mapsto (d_t, f_t^{(idx)}, \Tilde{x}_t)$ in Alg. 1.8-9. From the RGB image $\Tilde{x}_t$ we calculate the uncolored image mask $\Tilde{x}_m$ using HSV color space filtering of the background color (Alg. 1.10). These uncolored parts of the image are inpainted with the depth-aware inpainting ControlNet (Alg. 1.11) denoted as

\begin{equation}
    x_t = \mathcal F_\theta (\Tilde{x}_t, \Tilde{x}_m, g, d_t; \theta_i, \theta_a, \theta_c),
\end{equation}
where $\theta_i$ is the inpainting ControlNet, $\theta_a$ is the IP-Adapter encoder, $\theta_c$ is the depth ControlNet and $g$ is the appearance condition of a nearby webcam image and the text prompt. Then the inpainted image is textured onto a local UV texture map $\mathcal T_t: (M_{t-1}, x_t, p_t, f_t^{(idx)}, m_{t-1}^{UV}) \mapsto T_t$ and then superposed with the previous texture map as in equation~(\ref{eq:texturing}) and Alg. 1.12. After the loop we return the inpainted UV texture $T_I = T_N$ (Alg. 1.13), do a UV texture postprocessing to fill the remaining regions and save the inpainted frames and create a video. Or the inpainted images from several timestamps are used in section~\ref{sec:WildGaussians} to train a Wild Gaussian 3D scene.

\begin{algorithm}[H]
    \caption{SeasonScapes Training}
    \label{alg:seasonMesh}
    \begin{algorithmic}[1]
    \Require 
        \Statex \textbf{Inputs:}
        \begin{itemize}
            \item Webcam images $I_w$ and pose $p_w$
            \item Uncolored 3D mesh $M_0$ with UV texture $T_0$
            \item Inpainting path with poses $\{p_t\}_{t=1}^N$
            \item Diffusion parameters $\mathcal{F}_{\theta_d}$, $\mathcal{F}_{\theta_c}$, $\mathcal{F}_{\theta_i}$, $\mathcal{F}_{\theta_a}$
            \item Text, IP-Image guidance $g$
        \end{itemize}
    \Ensure Estimate the UV texture $T$
    \For{$t= 1$ to $W$}
        \State{\# Paint}
        \State{Render valid face indices: $f_t^{(idx)}  \leftarrow \mathcal R(M, p_t)$}
        \State{Texture: $T_t \leftarrow \mathcal T(M_{t-1}, I_t, f_t^{(idx)}, m_{t-1}^{UV})$}
    \EndFor
    \State{$T_0 \leftarrow T_W$, $M_0 \leftarrow M_W$}
    \For{$t=1$ to $N$}
        \State{\# Inpaint}
        \State{Render depth, face indices: $d_t, f_t^{(idx)}  \leftarrow \mathcal R(M, p_t)$}
        \State{Render RGB: $\Tilde{x}_t \leftarrow \mathcal R(M_{t-1}, p_t)$}
        \State{Mask RGB: $\Tilde{x}_m \leftarrow \text{Mask}(\Tilde{x}_t)$}
        \State{Inpaint: $x_t \leftarrow \mathcal F_\theta(\Tilde{x}_t, \Tilde{x}_m, g, d_t; \theta_i, \theta_a, \theta_c)$}
        \State{Texture: $T_t \leftarrow \mathcal T(M_{t-1}, x_t, f_t^{(idx)}, m_{t-1}^{UV})$}
    \EndFor
    \State \Return $T_I$
    \end{algorithmic}
\end{algorithm}

\subsection{Time Embedding using Wild Gaussian}
\label{sec:WildGaussians}
After performing inpainting on the sparse input views, we obtain denser camera views along the trajectory. By repeating this process across multiple timestamps, we generate images with both spatial and temporal variations. These images, combined with 3D mesh, are then used for training relightable Gaussians within the wWild Gaussian framework with learned time embeddings as stated in \citet{kulhanek2024wildgaussians3dgaussiansplatting}. The color attributes of each 3D Gaussian, denoted as $\hat{c}_i(r)$, are transformed via an affine operation parameterized by an appearance MLP.  $\tilde c_i = \gamma \cdot \hat c_i(r) + \beta.$ The MLP outputs coefficients $(\beta, \gamma) \in (\mathbb{R}^3, \mathbb{R}^3)$ conditioned on the viewing ray direction.  $r$.

\begin{figure*}[t]
    \begin{minipage}[c]{0.04\linewidth}
        \subcaption{}
    \end{minipage}
    \begin{minipage}[c]{0.96\linewidth}
        \includegraphics[width=0.32\linewidth]{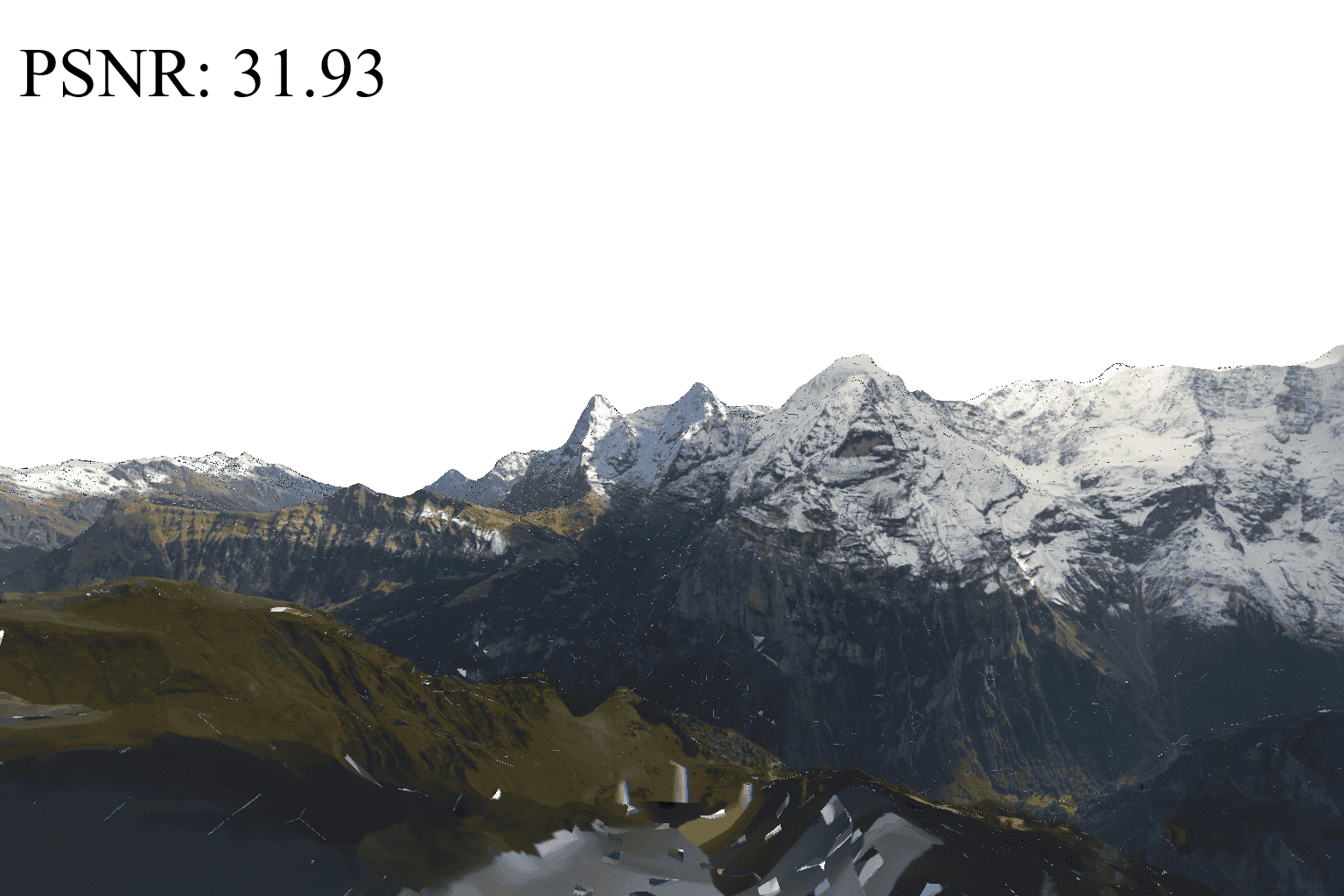}
        \includegraphics[width=0.32\linewidth]{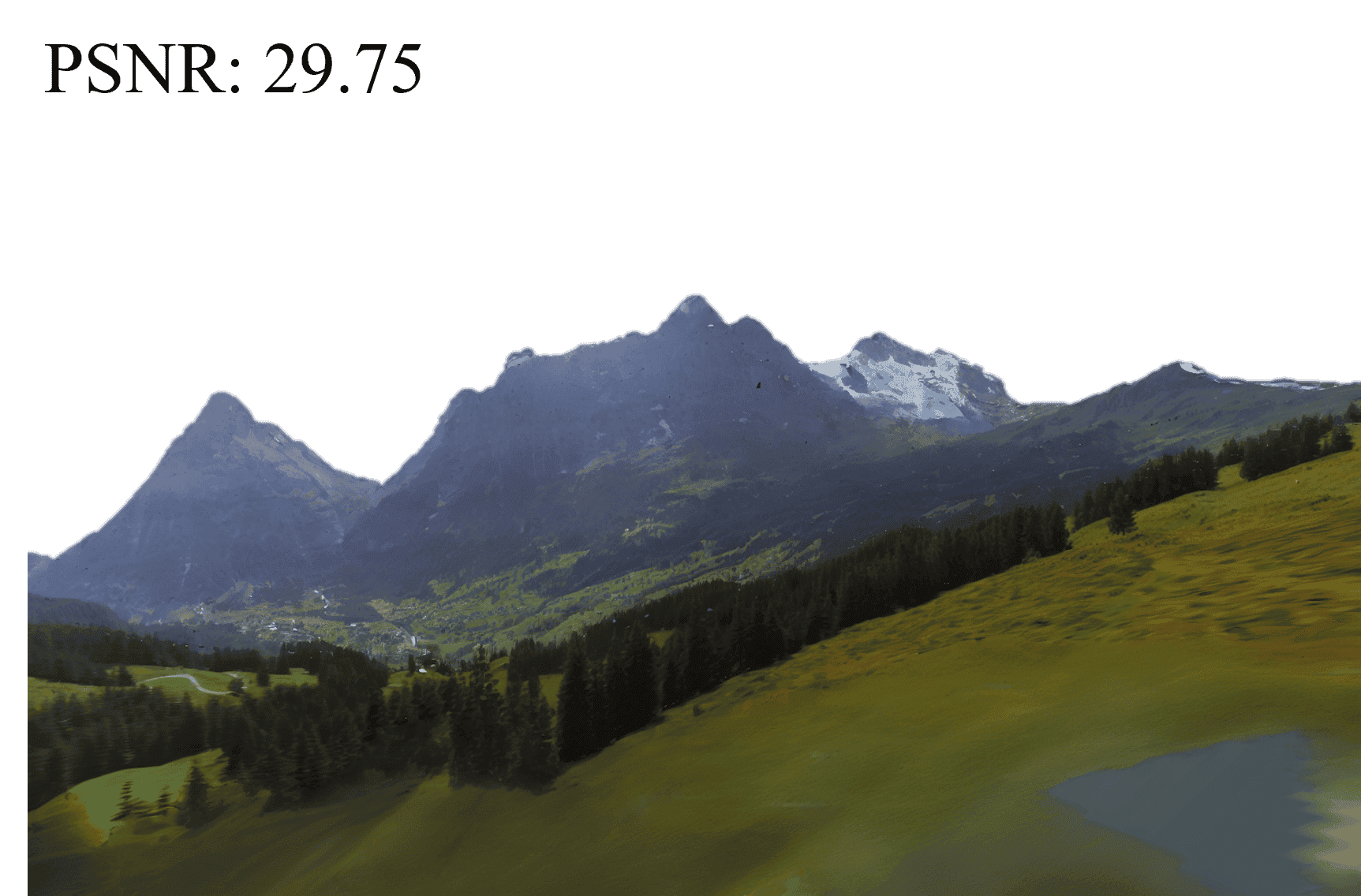}
        \includegraphics[width=0.32\linewidth]{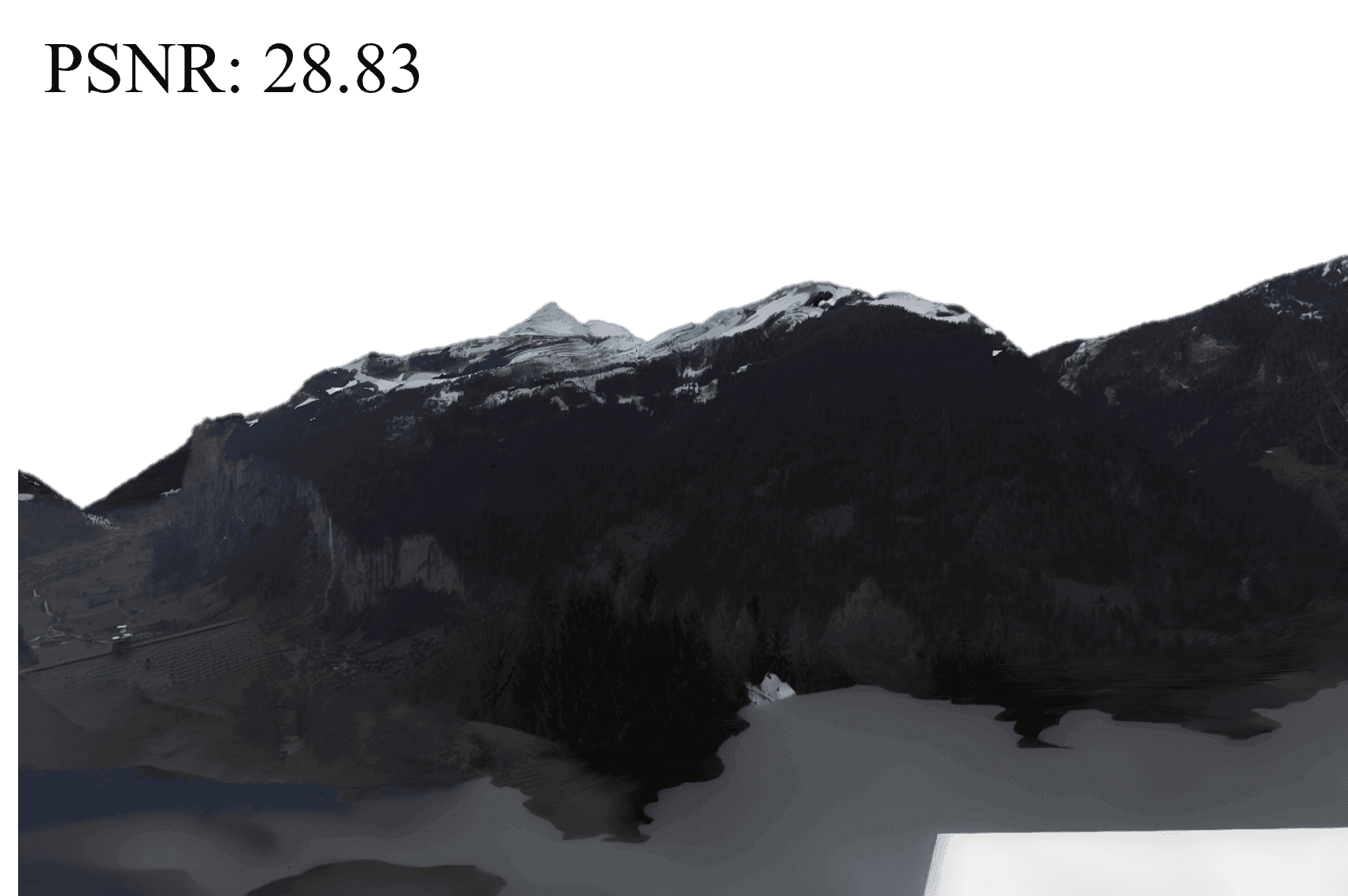}    
    \end{minipage}
    \begin{minipage}[c]{0.04\linewidth}
        \subcaption{}
    \end{minipage}
    \begin{minipage}[c]{0.96\linewidth}
        \includegraphics[width=0.32\linewidth]{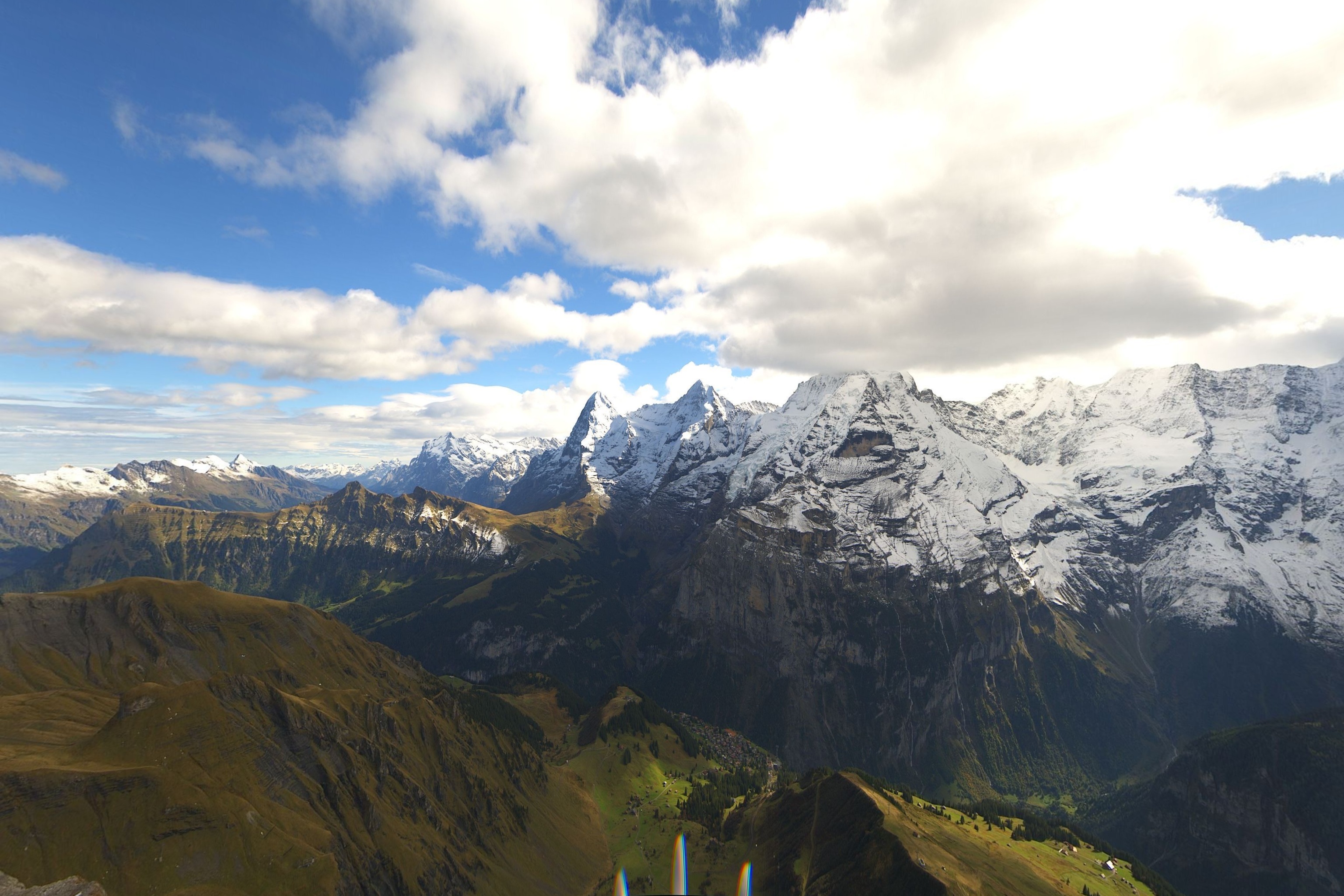}
        \includegraphics[width=0.32\linewidth]{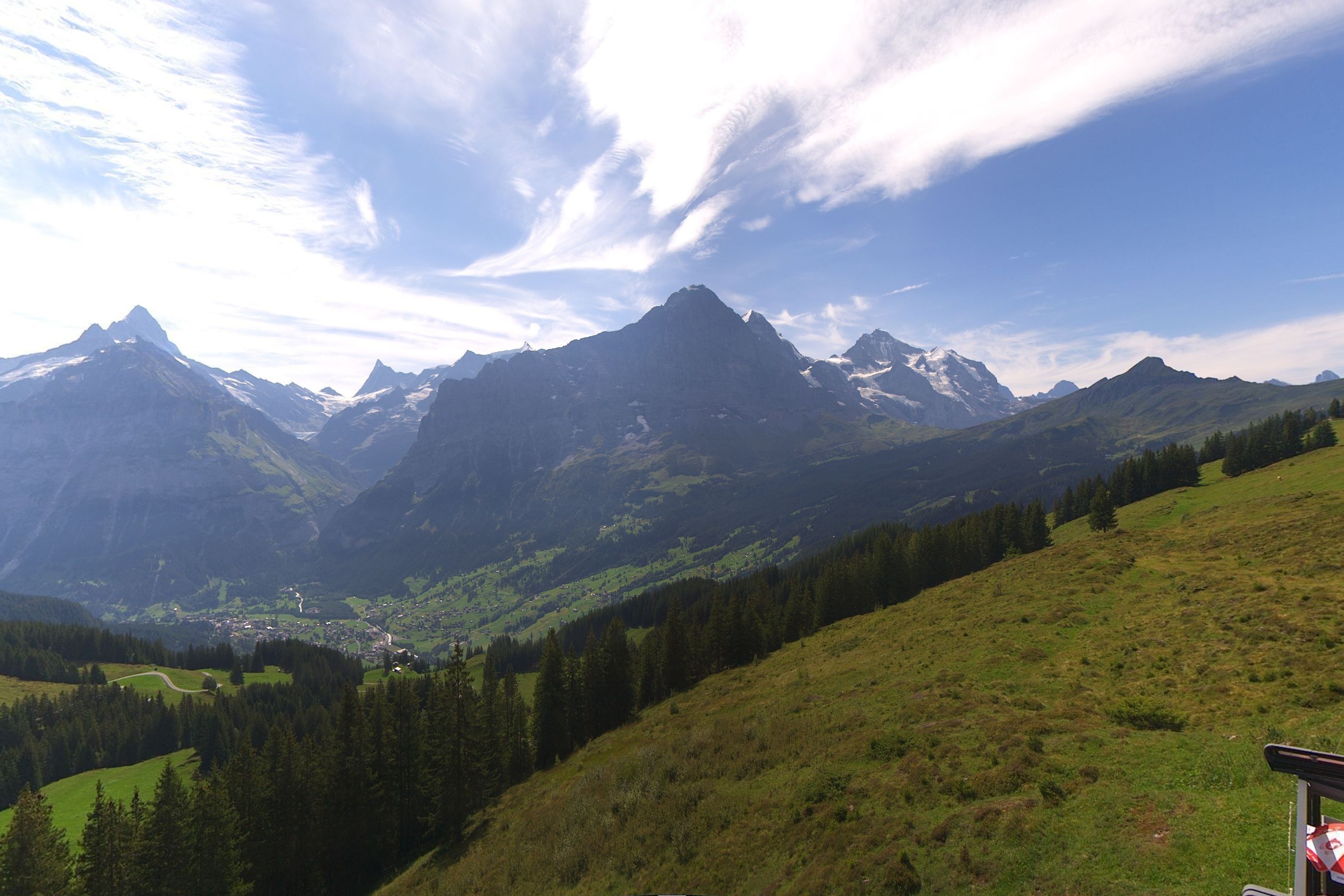}
        \includegraphics[width=0.32\linewidth]{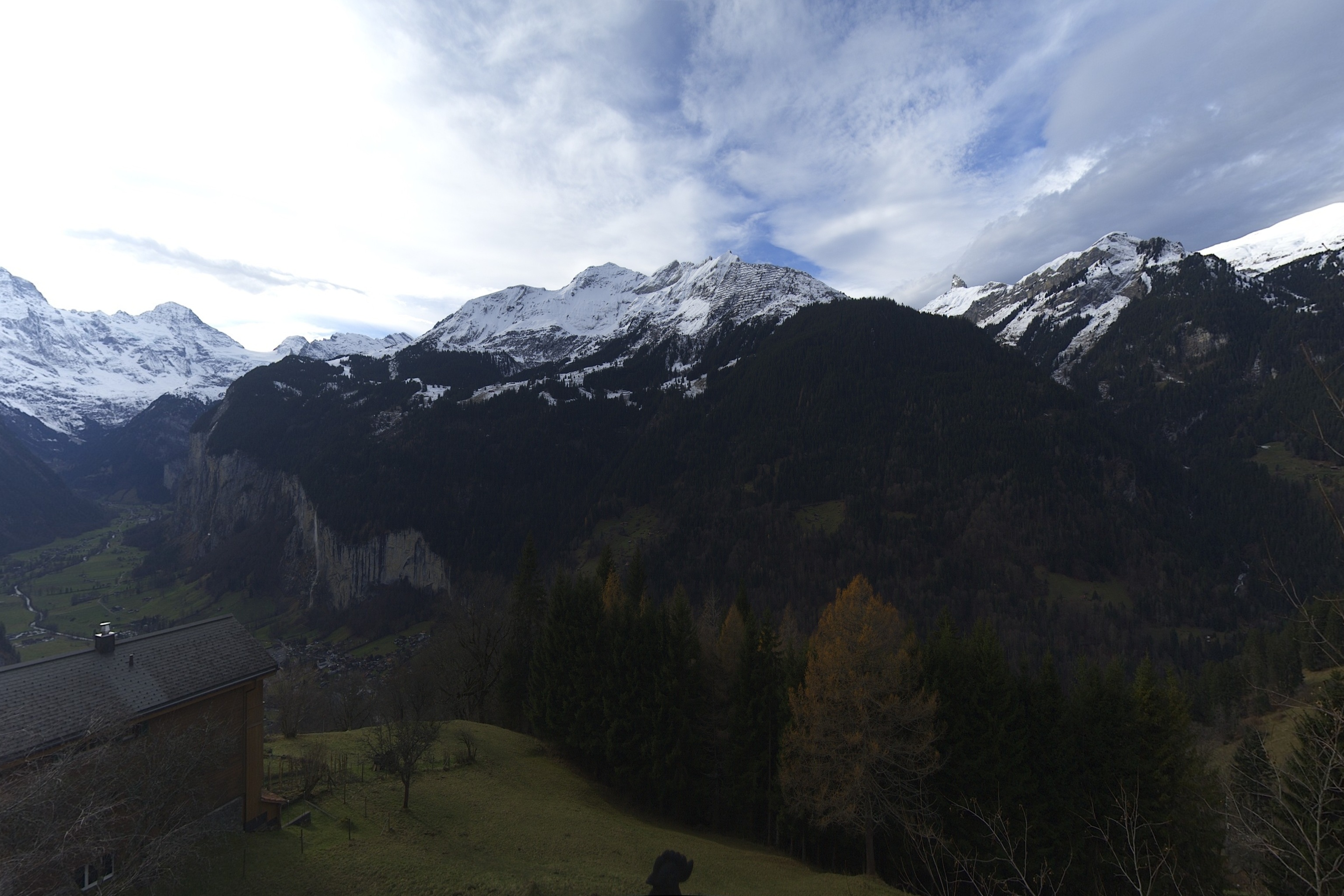}
    \end{minipage}
    \caption{\textbf{Qualitative comparison.} (a) shows the inpainted images from our 3D landscape mesh without the clouds and in row (b) are the ground truth webcam images. The evaluation is done without the sky and without the nearest parts of the mesh. The first image is from the 7.10.2024 at 12 AM, the second image is from the 20.7.2024 at 12 AM, third image is from the 7.12.2024 at 12 AM. }
    \label{fig:painting_jungfrauRegion}
\end{figure*}

\section{Experiments}
\label{sec:experiments}

To evaluate our SeasonScapes pipeline, we present both qualitative comparisons against Google Earth's visual outputs and quantitative metrics of our pipeline. Firstly, we introduce the dataset settings and evaluation metrics. Then we conduct an ablation study to demonstrate the effectiveness of each module in our framework.

\noindent{\textbf{Evaluation Procedure.}} We evaluate the inpainted images with commonly used metrics for reconstruction quality. Specifically we report the the Learned Perceptual Image Patch Similarity (LPIPS) with AlexNet~\cite{LPIPS_score}, Peak-Signal-to-Noise Ratio (PSNR) and Structural Similarity Index Measure (SSIM)~\cite{ssmi_score}. We prioritize the LPIPS metric for evaluation because our test views combine visible regions from the initial viewpoint with unseen areas requiring plausible appearance synthesis. Unlike pixel-level metrics that fail to assess generated content meaningfully, LPIPS effectively measures perceptual similarity for real and synthesized regions, ideal for mixed-scenario validation. To calculate the score we render inpainted images in the resolution of $1024\times 1536$ at the webcam location and compare them to the actual webcam. As the sky with the clouds are artificially generated we mask out the sky and the near region as the webcams often has houses here.

\subsection{Quantitative and Qualitative Evaluation}

We evaluate our 3D landscape texturing in 3 different areas of the 3D mesh. The first Grindelwald scene is a small scene with one mountain that is viewed from different directions. Whereas the Jungfrau scene consists of several mountains and valleys and is painted with 14 webcam images. The last scene consists of a large part of the Bernese highlands with 21\,km $\times$ 36.5\,km. In case of the Grindelwald scene we average 18 different timestamps and we get a LPIPS value of 0.159. In case of the larger Jungfrau scene with 14 different timestamps this value increase to 0.191, as seen in Tab.~\ref{tab:painting_metrics} and for the Bernese highland scene with 4 different timestamps we get a LPIPS of 0.216. The performance degradation is primarily attribute to increased unseen regions in evaluation views. We provide detailed analysis and visual examples demonstrating the quality variation between optimally and poorly rendered regions.


\begin{table}[!htb]
    \centering
    \caption{\textbf{Quantitative evaluation.} We list a small scene, Grindelwald, which consists of 2 painting views and 1 evaluation view per timestamp and a larger scene, Jungfrau region, that has 14 painting views and 41 evaluation views for each timestamp. The number of inpainting view $N$ appears in parentheses after each scene name.}
    \resizebox{0.85\linewidth}{!}{
    \begin{tabular}{p{2.6cm}ccc}
    \thickhline
    \textbf{Region}  & \textbf{LPIPS} $\downarrow$ & \textbf{PSNR} $\uparrow$ & \textbf{SSIM} $\uparrow$ \\
    \hline
    \textbf{Grindelwald} (54) & \textbf{0.159}  & \textbf{20.56} & \textbf{0.86}\\
    \textbf{Jungfrau} (674) & 0.191  & 18.58  & 0.831\\
    \textbf{Bern high} (168) & 0.216 & 19.20  & 0.804 \\
    \thickhline
    \end{tabular}
    }
    \label{tab:painting_metrics}
\end{table}

We qualitatively compare the ground truth webcam images and our inpainted renderings in Fig.~\ref{fig:painting_jungfrauRegion}. To ensure a rigorous evaluation of topographic reconstruction, we apply a binary mask to the sky and foreground regions, thereby isolating the mountain facets and mitigating the impact of painting-related artifacts in the near-field from skewing the quantitative results. on the PSNR calculation. Results demonstrate consistent artifacts in region which is fully painted by ControlNet, like the valley region in first column, the grassy area and the forested zone in second and third column.

\subsection{Ablation Studies}
\label{sec:ablation}
\begin{figure}[h]
    \centering
    \begin{minipage}[c]{0.32\linewidth}
        \includegraphics[width=1\linewidth]{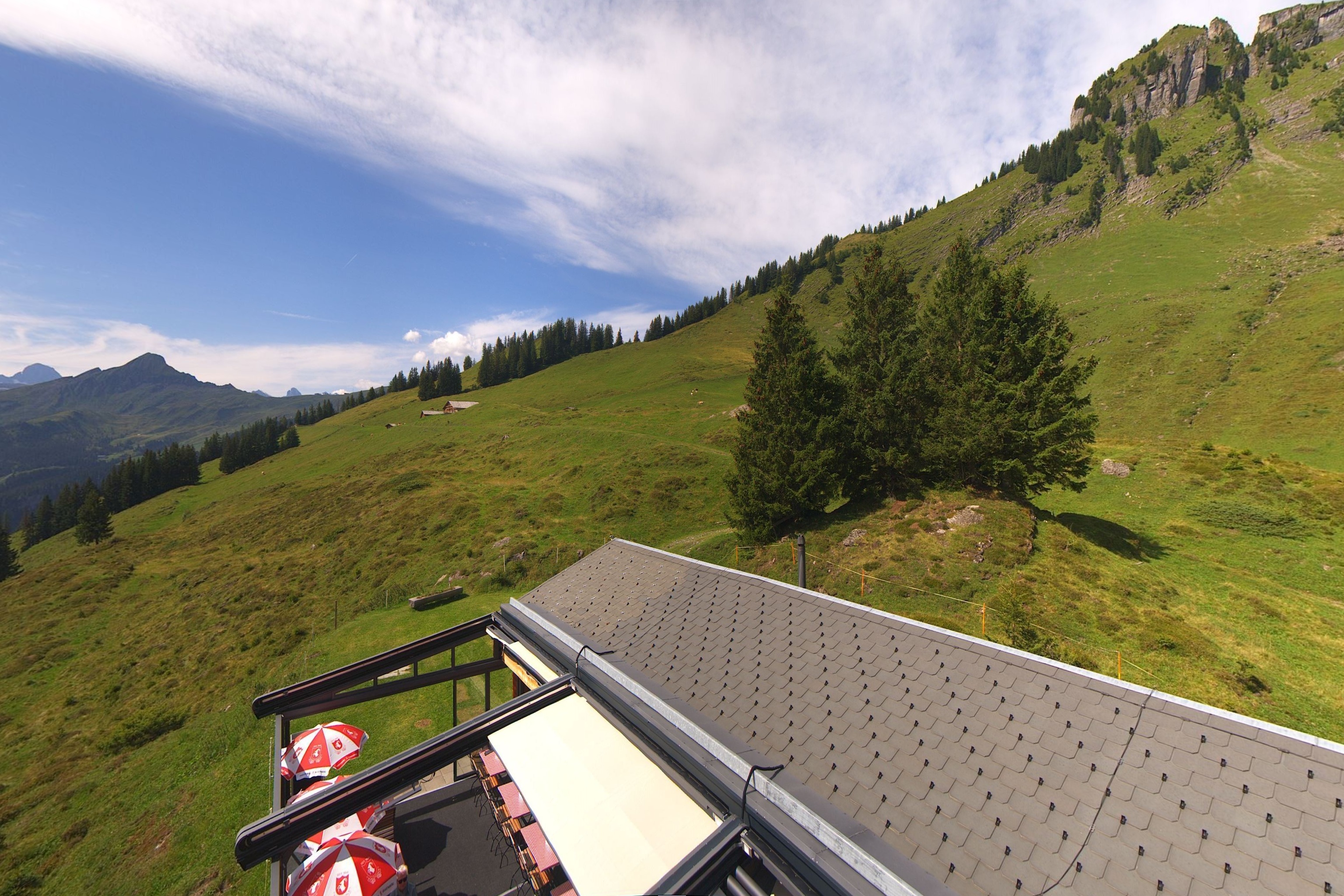}
        \subcaption[c]{}
    \end{minipage}
    \begin{minipage}[c]{0.32\linewidth}
        \includegraphics[width=1\linewidth]{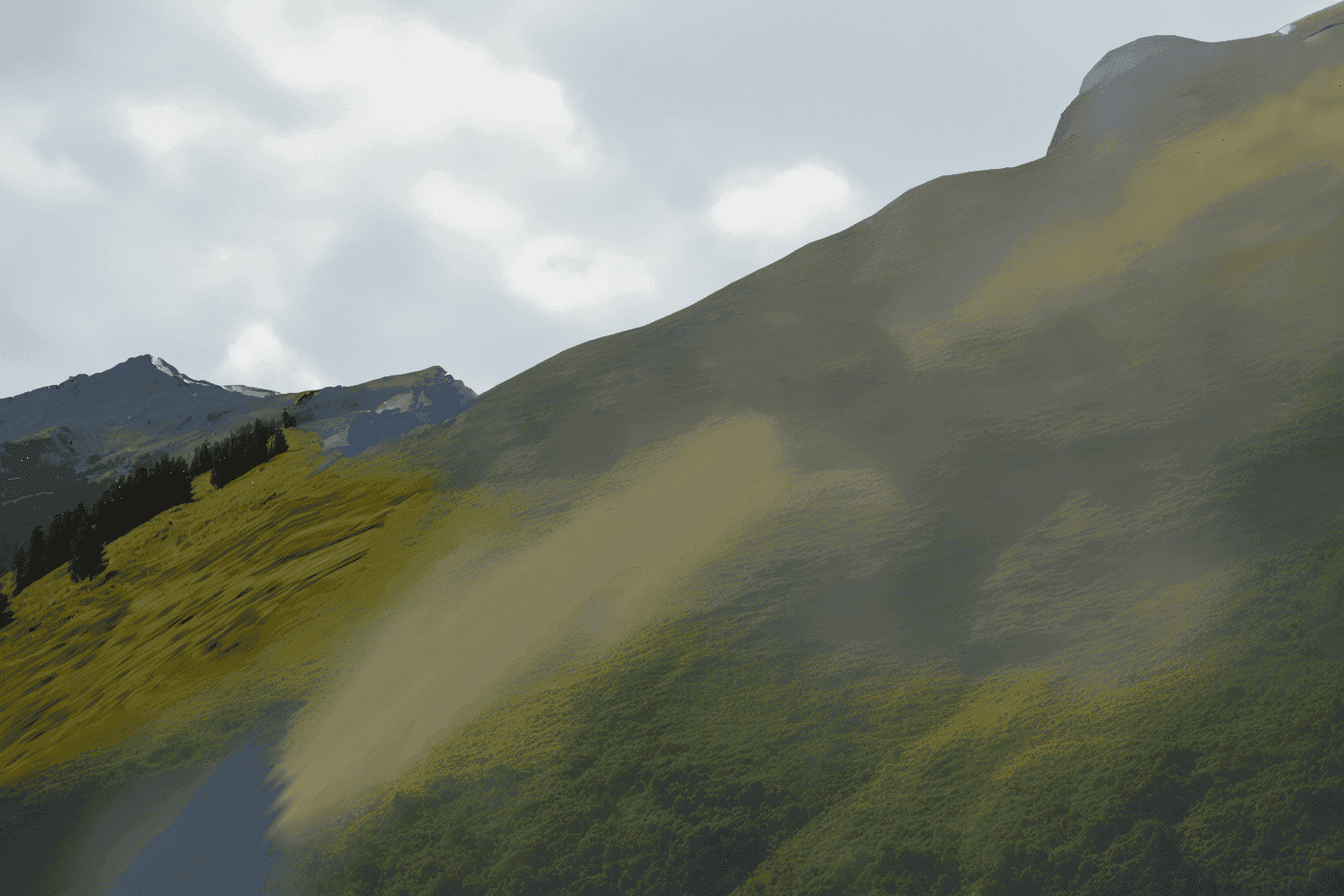}
        \subcaption[c]{}
    \end{minipage}
    \begin{minipage}[c]{0.32\linewidth}
        \includegraphics[width=1\linewidth]{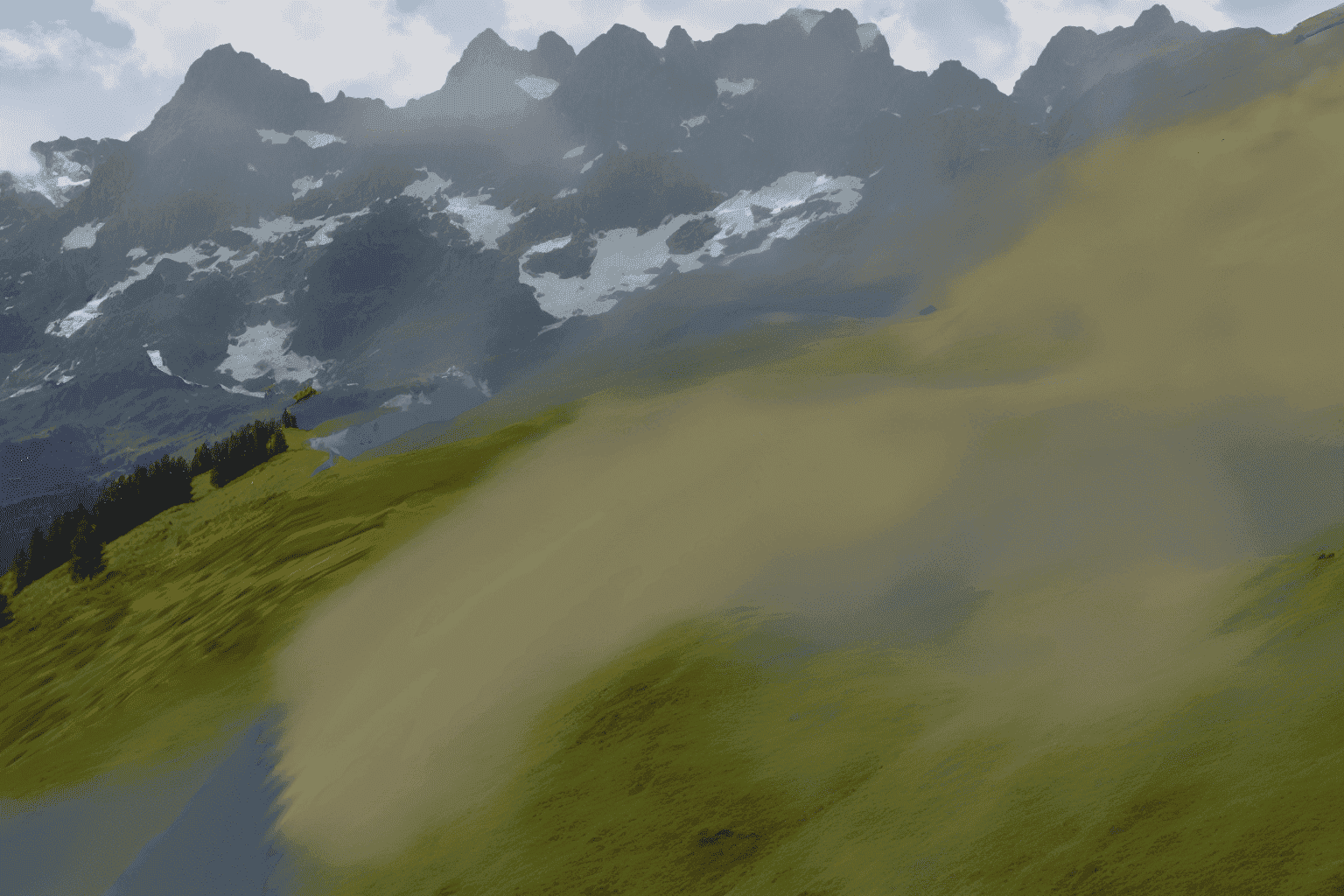}
        \subcaption[c]{}
    \end{minipage}
    \begin{minipage}[c]{0.32\linewidth}
        \includegraphics[width=1\linewidth]{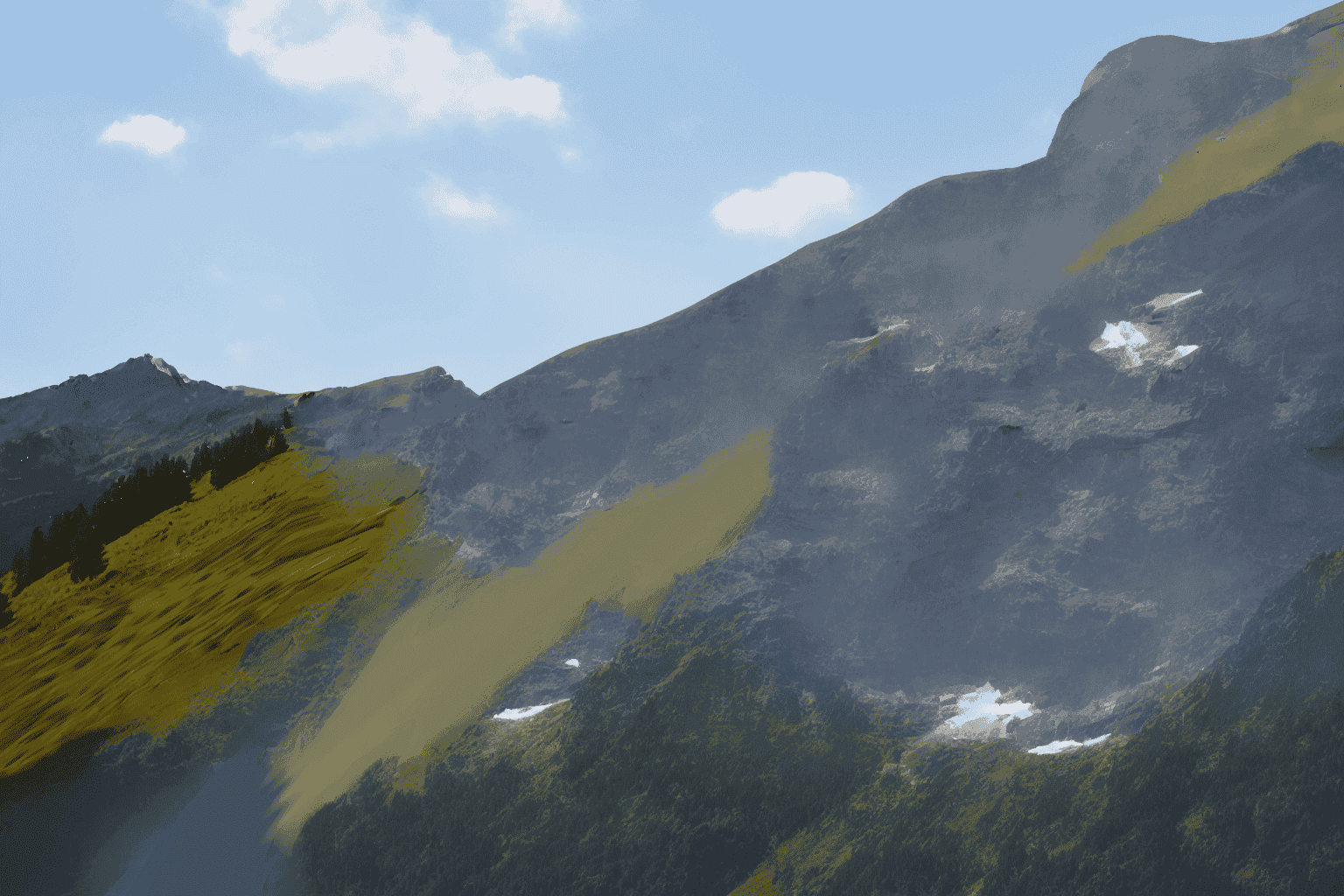}
        \subcaption[c]{}
    \end{minipage}
    \begin{minipage}[c]{0.32\linewidth}
        \includegraphics[width=1\linewidth]{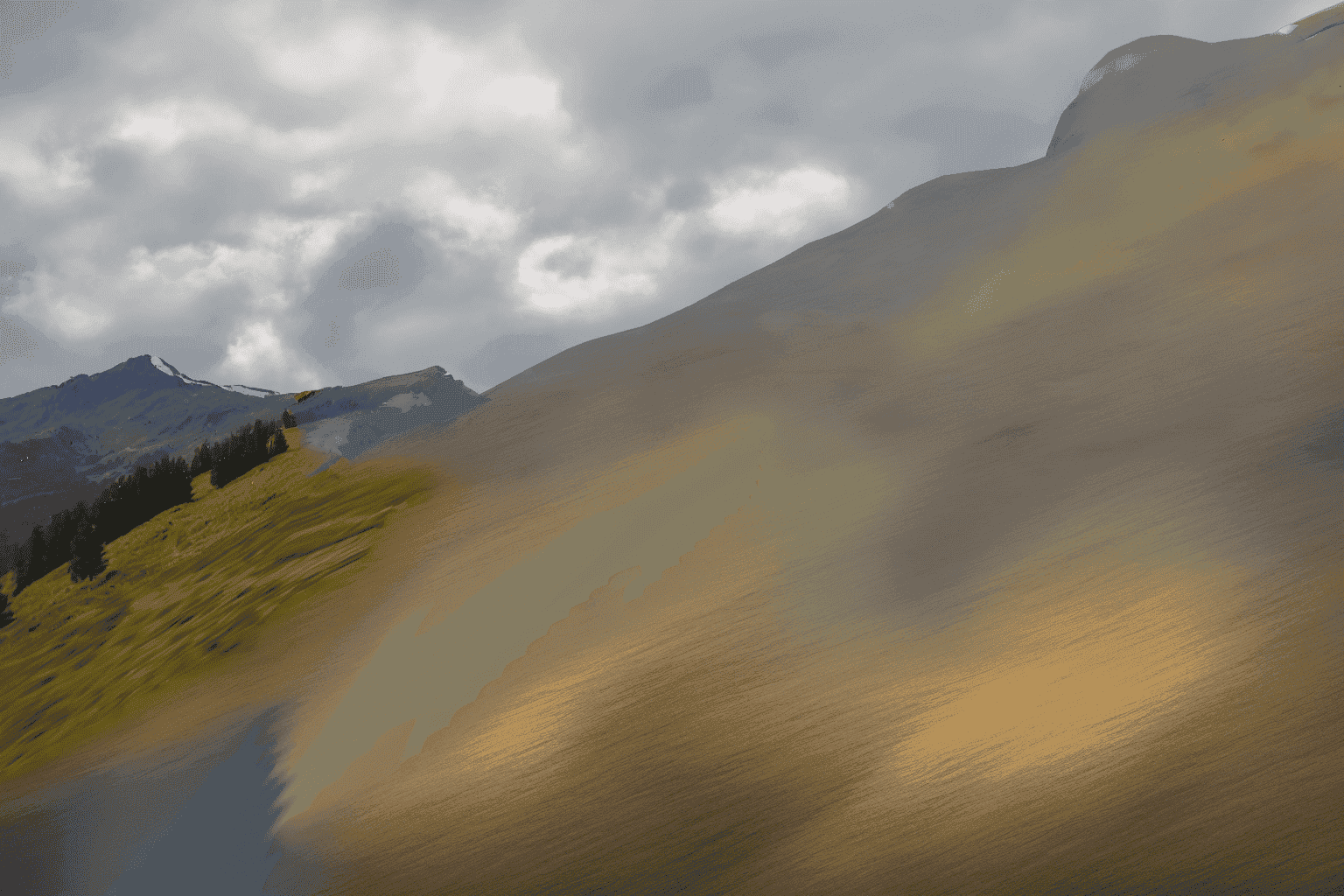}
        \subcaption[c]{}
    \end{minipage}
    \begin{minipage}[c]{0.32\linewidth}
        \includegraphics[width=1\linewidth]{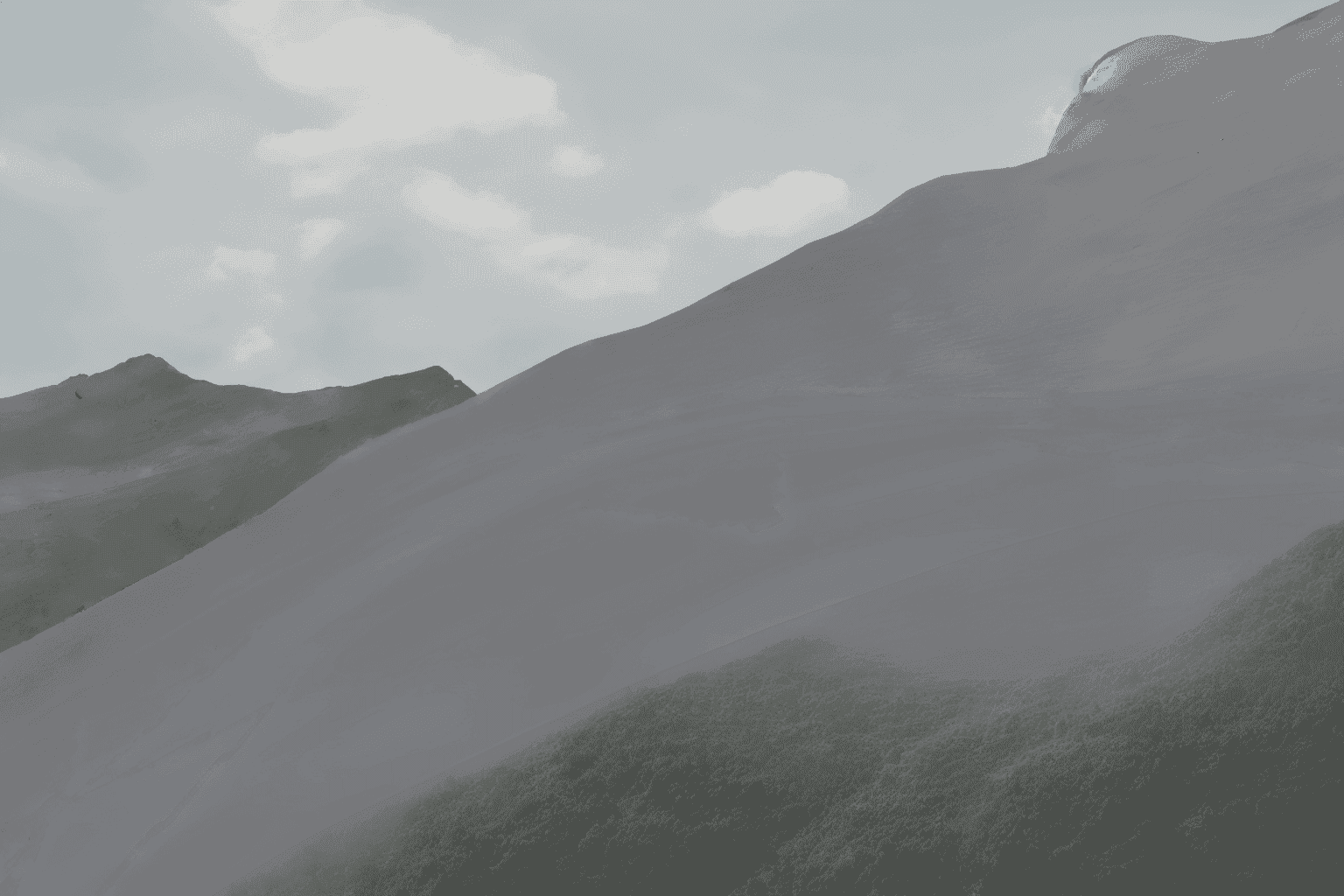}
        \subcaption[c]{}
    \end{minipage}
    \caption{\textbf{Qualitative comparison of ablation study.} We evaluate six configurations: (a) target webcam, (b) with all condition, (c) without depth condition, (d) without inpainting mask condition, (e) without IP-Adapter image condition and (f) no initial painting stage. }
    \label{fig:painting_ablation}
\end{figure}

We perform an ablation study of our SeasonScapes pipeline, systematically removing components and analyzing their impact through quantitative metrics in Tab.~\ref{tab:painting_metrics_ablation} and qualitatively in Fig.~\ref{fig:painting_ablation}. 


\begin{table}[h]
    \centering
    \caption{\textbf{Quantitative results of the ablation study.} We evaluate how different control conditions affect performance across both the smaller scale Grindelwald scene and larger scale Jungfrau region, with detailed qualitative explanation shown in  \cref{fig:painting_ablation}.}
    \vspace{1em}
    \resizebox{0.95\linewidth}{!}{
    \begin{tabular}{lcccc}
    \thickhline
    \textbf{Ablation study}  & \textbf{LPIPS}$\downarrow$ & \textbf{PSNR}$\uparrow$ & \textbf{SSIM}$\uparrow$ & \textbf{Drop} \\
    \hline
    \multicolumn{5}{l}{\textbf{Grindelwald 1. September 2024}} \\
    All & \textbf{0.145} & \textbf{23.02} & \textbf{0.889} &   \\        
    W/o prompt&0.149&22.92&0.886& 2.8\,\%  \\
    W/o depth cNet & 0.150 & 22.56 & 0.884 & 3.5\,\%  \\
    W/o inpaint mask & 0.152& 22.43 & 0.882 & 4.8\,\%  \\
    W/o IP-Adapter& 0.148 & 22.03 & 0.880 & 2.0\,\%  \\
    W/o paint& 0.193& 19.40 & 0.863  & 33.1\,\% \\
    \hline
    \multicolumn{5}{l}{\textbf{Jungfrau region 1. September 2024}} \\
    All& 0.184 & 19.65 & \textbf{0.833}  & \\        
    W/o prompt& \textbf{0.183} & \textbf{19.66}  & 0.832  & -0.5\,\% \\
    W/o depth cNet& 0.185 & 19.57 & 0.830  & 0.5\,\% \\
    W/o inpaint mask& \textbf{0.183} & 19.38 & 0.829  & -0.5\,\%  \\
    W/o IP-Adapter& 0.184 & 19.65 & 0.832  & 0.0\,\% \\
    W/o paint& 0.206 & 18.02 & 0.830  & 12.0\,\%   \\
    \thickhline
    \end{tabular}
    }
    \label{tab:painting_metrics_ablation}
\end{table}

We observe a significant performance drop when skipping the first-stage painting initialization in both scenes. Note that the performance drop is more pronounced in the Grindelwald scene compared to the Jungfrau region. We attribute this to the Jungfrau's primary bottleneck being its extensive unseen areas and overall lower quality - where synthesized content dominates and control signals have reduced impact. This observation is supported by comparable LPIPS results despite noticeable pixel-level performance drops. In Fig.~\ref{fig:painting_ablation} the result demonstrate that omitting depth control leads to mountain hallucinations in distant regions and the IP-Adaptor mainly primarily modulates color tones in the synthesized image.

\subsection{Qualitative Comparison with Google Earth}

\begin{figure*}[!t]
    \centering
    \includegraphics[width=1\linewidth]{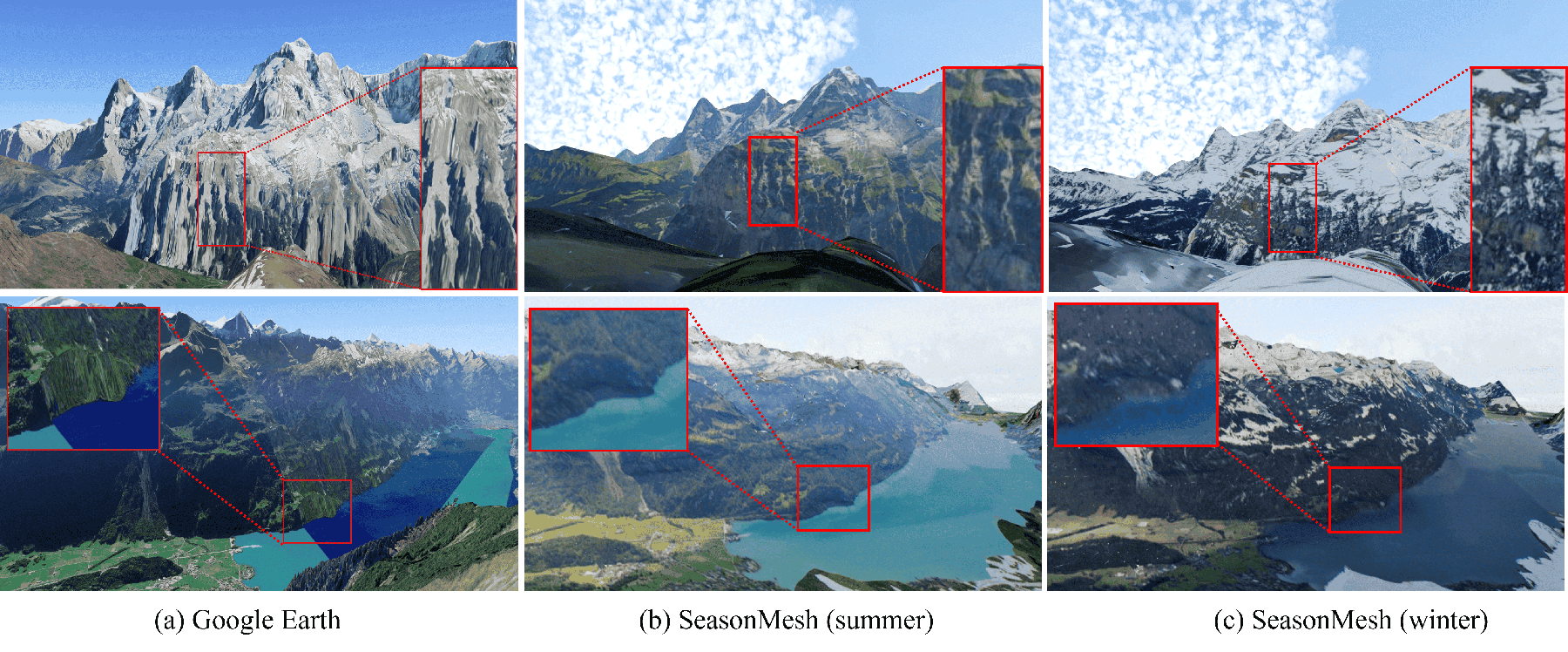}
   \caption{\textbf{Qualitative comparison between our approach and Google Earth.} Images (a) are from Google Earth \cite{GoogleEarth}, images (b) are our SeasonScapes rendering from the 1.9.2024 at 12 AM, images (c) are our SeasonScapes from the 18.12.2025 at 10 AM. SeasonScapes shows superior photorealistic rendering and time variant ability}
    \label{fig:Google_earth_compare}
\end{figure*}

In Fig.~\ref{fig:Google_earth_compare} we compare our results with rendered 3D Landscape from Google Earth. Our method demonstrates superior coloring of steep cliff faces by better capturing near-orthogonal views, in contrast to Google Earth’s satellite-based perspective, which often misses details on vertical surfaces due to its top-down viewpoint and warping. Furthermore, our method adapts more efficiently to seasonal variations due to reduced weather dependency. These results suggest that our approach effectively synthesizes high-fidelity textures especially for vertical topographies.

\subsection{Wild Gaussian}
\begin{figure}[h]
    \begin{minipage}[c]{0.04\linewidth}
        \subcaption{}
    \end{minipage}
    \begin{minipage}[c]{0.96\linewidth}
        \includegraphics[width=0.32\linewidth]{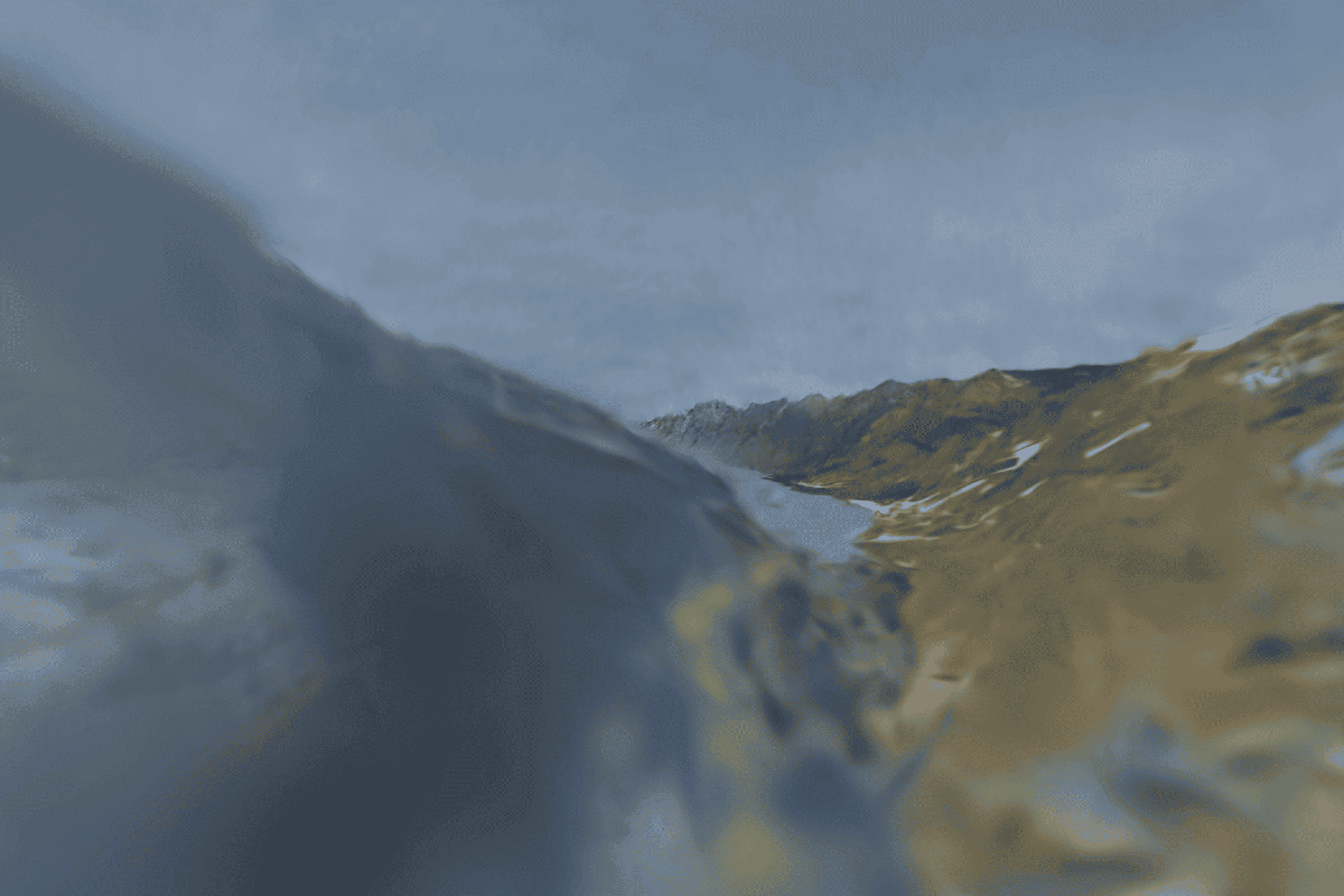}
        \includegraphics[width=0.32\linewidth]{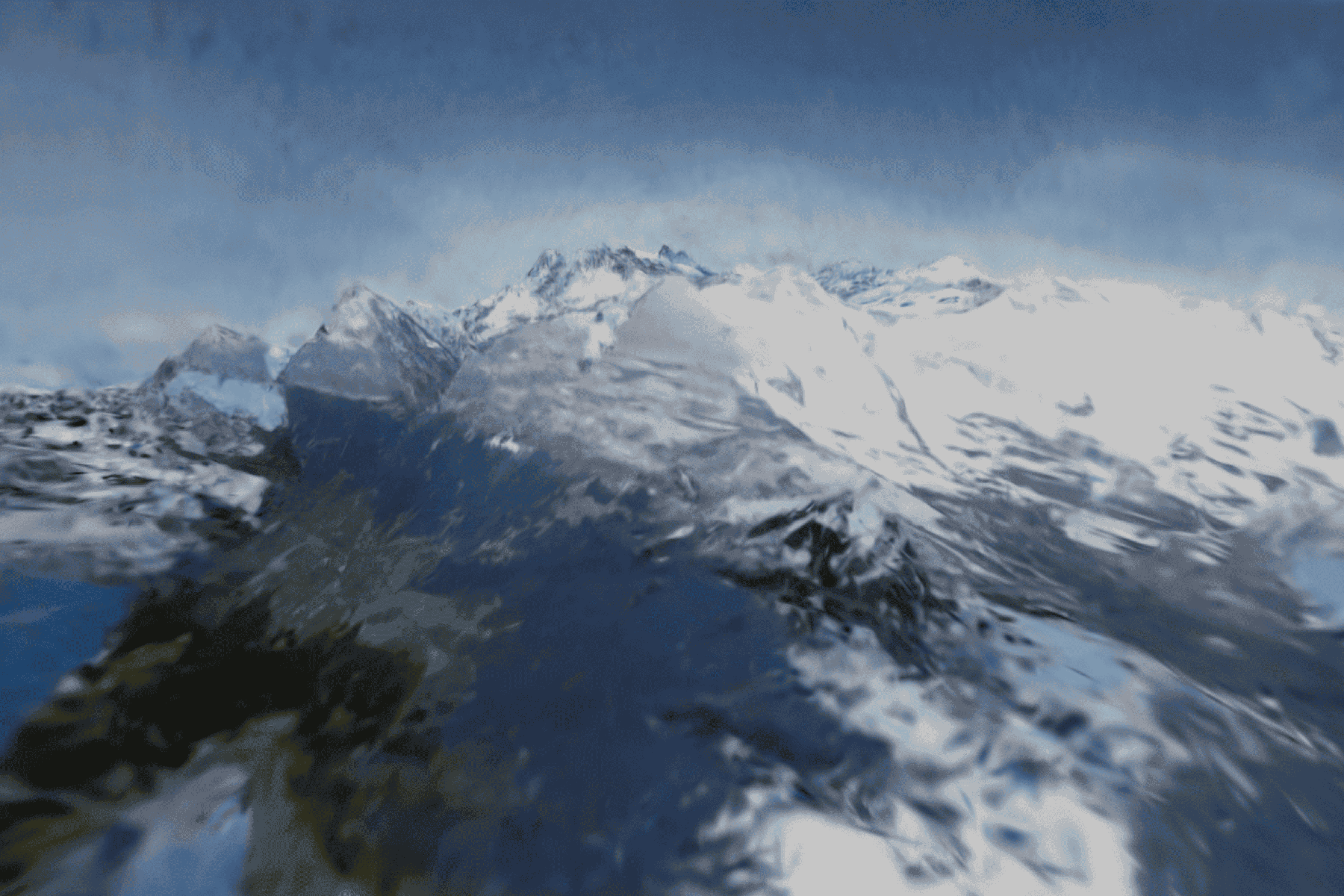}
        \includegraphics[width=0.32\linewidth]{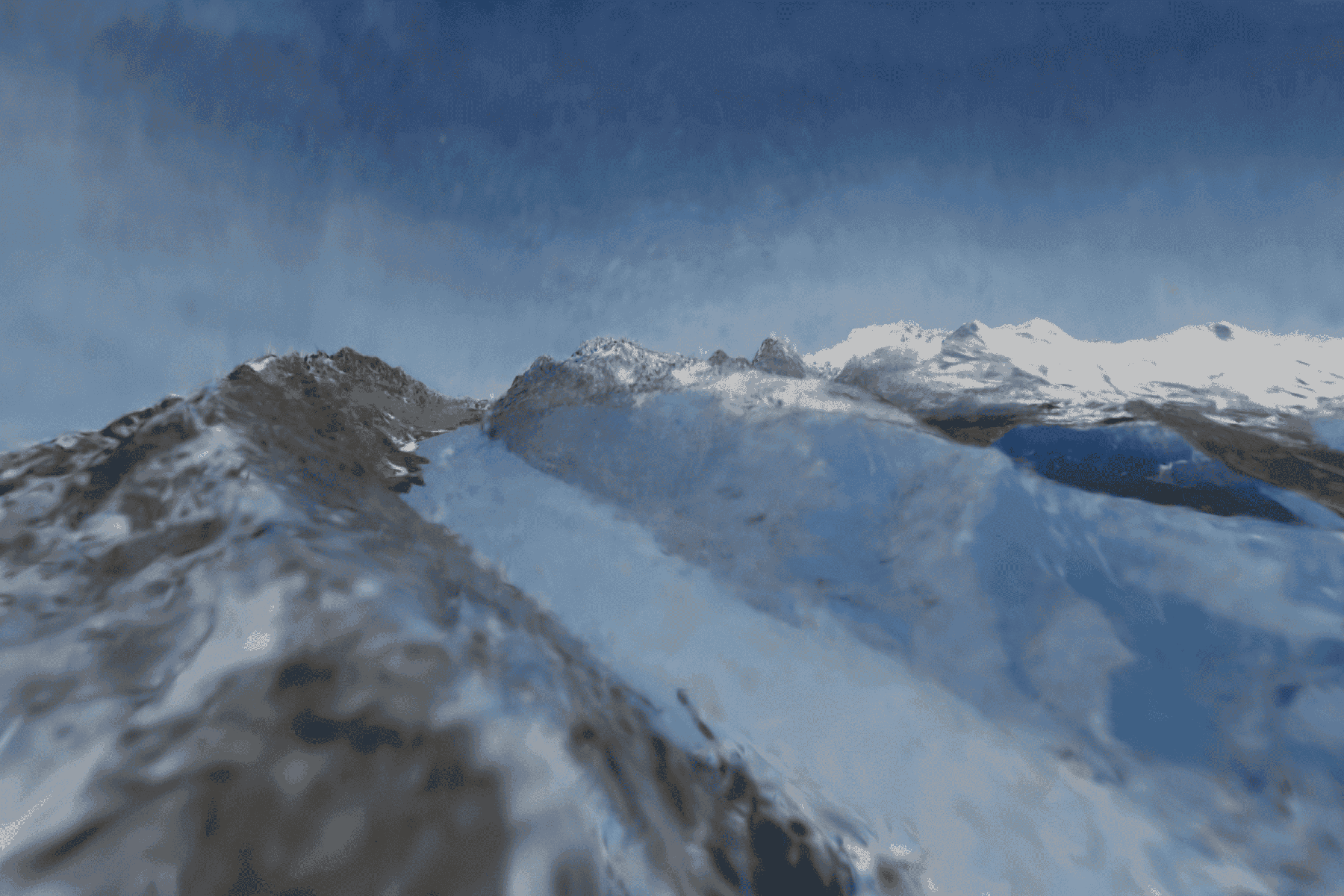}    
    \end{minipage}
    \begin{minipage}[c]{0.04\linewidth}
        \subcaption{}
    \end{minipage}
    \begin{minipage}[c]{0.96\linewidth}
        \includegraphics[width=0.32\linewidth]{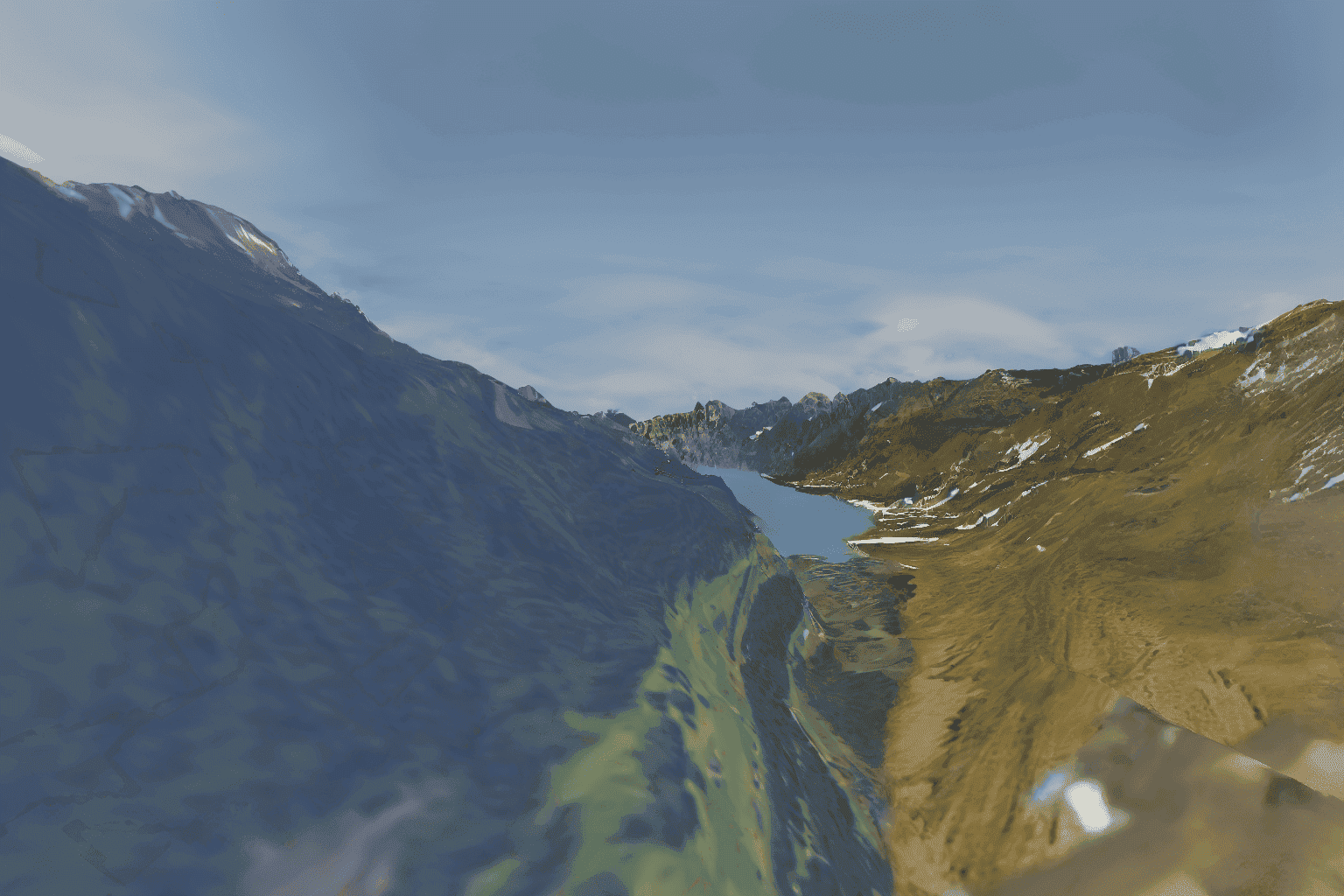}
        \includegraphics[width=0.32\linewidth]{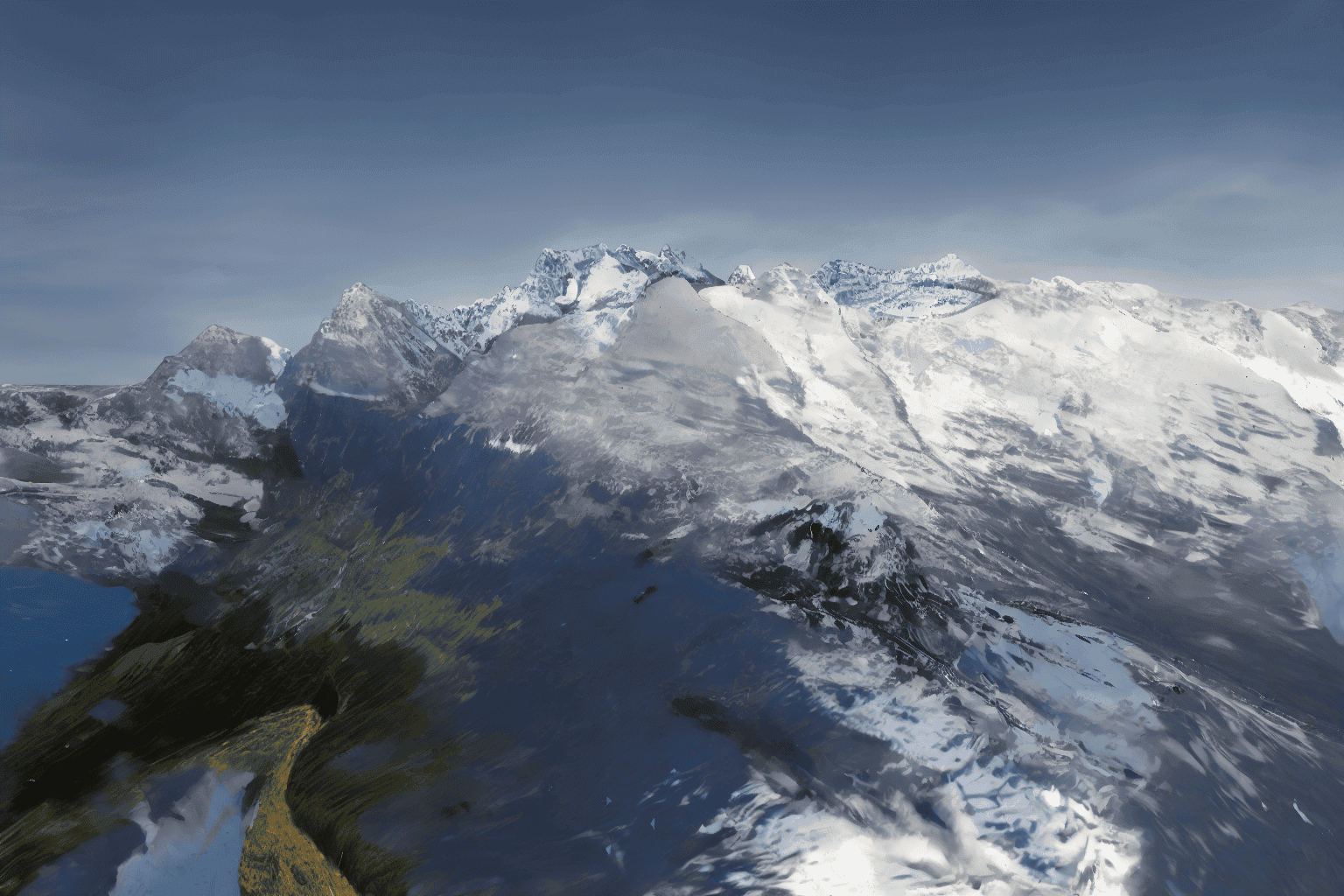}
        \includegraphics[width=0.32\linewidth]{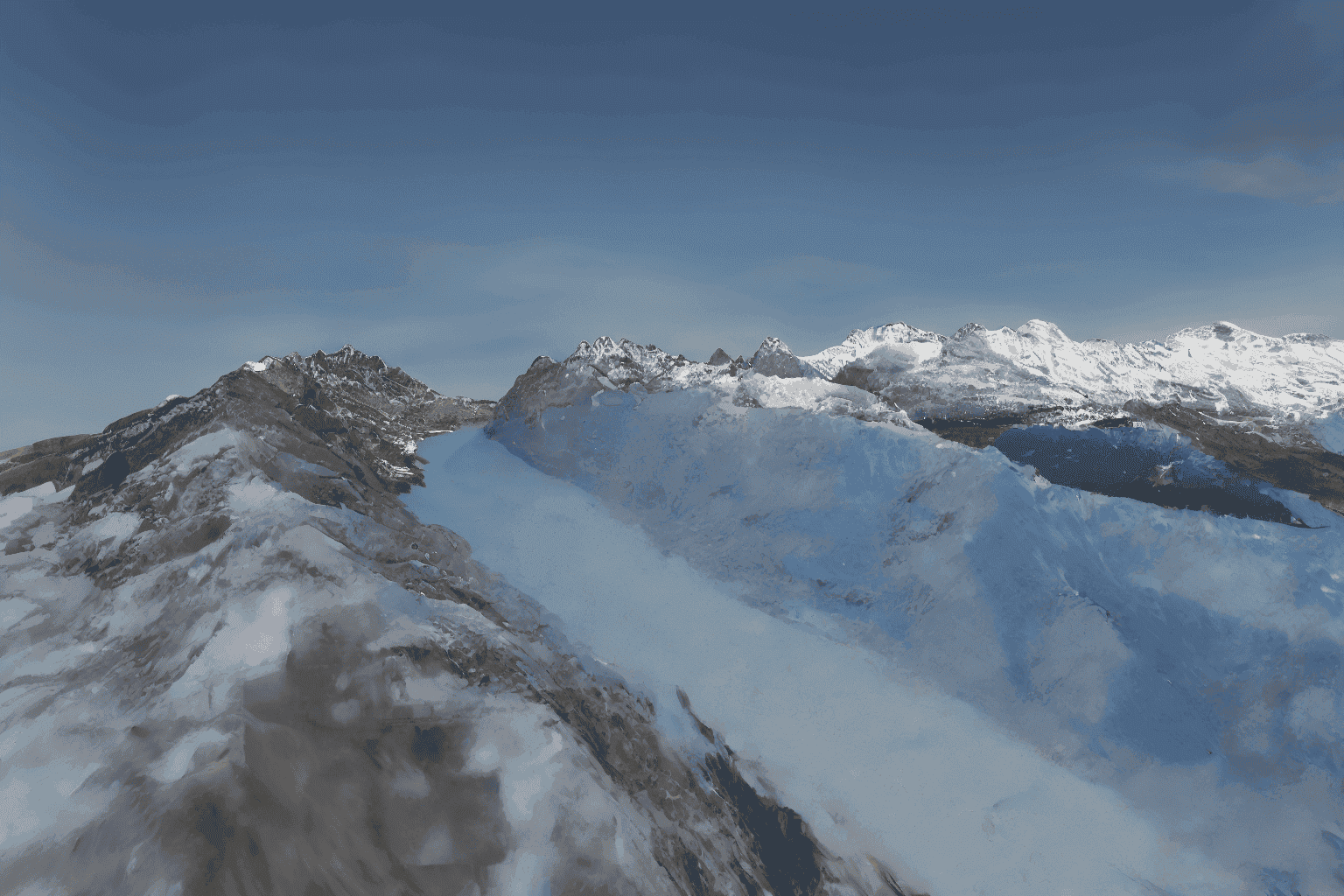}
    \end{minipage}
    \caption{\textbf{Qualitative comparison of novel Wild Gaussian views.} (a) Rendered relightable Gaussian outputs and (b) are the inpainted images from our iterative inpainting result. The first and the last image is from the 18.5.2025 at 2 PM, the second image is from the 12.5.2024 at 12 AM. }
    \label{fig:wG_novel}
\end{figure}

In Fig.~\ref{fig:wG_novel} we show the results of novel-view using wild Gaussian. Since our relightable Gaussian model is trained using synthesized ControlNet outputs along a trajectory, we primarily evaluate qualitative performance. The rendered images successfully capture both the structural and appearance variations across different timestamps.

\section{Conclusion}
\label{sec:conclusion}

We introduce SeasonScapes dataset, a 3D landscape dataset and the corresponding painting pipeline that textures meshes using webcam images while iteratively inpainting missing regions along novel camera trajectories through inpainting ControlNet. The approach robustly adapts to seasonal and diurnal variations by combining multi-source inputs - webcam images, low-resolution digital elevation models (DEMs), and satellite imagery - with camera parameter optimization. We additionally train a relightable Gaussian model using trajectory views, demonstrating robust performance in capturing diverse lighting conditions and appearance variations.



Our approach currently faces several limitations. The manual 2D-3D correspondence process remains time-consuming, and we observe unsatisfactory performance in ControlNet's ability to inpaint small missing regions. Additionally, the relightable Gaussian outputs occasionally exhibit blurry artifacts, likely due to calibration errors and inherent expressiveness constraints of the Wild Gaussian framework. These limitations present clear directions for future improvement.

\clearpage

\noindent{\textbf{Acknowledgments}} \\
This research was partially funded by the Ministry of Education and Science of Bulgaria (support for INSAIT, part of the Bulgarian National Roadmap for Research Infrastructure). We would like to thank the company Roundshot for providing the panoramic webcam images. 

{
    \small
    \bibliographystyle{ieeenat_fullname}
    \bibliography{main}

@String(CVPR= {IEEE Conf. Comput. Vis. Pattern Recog.})

@String(ICCV= {Int. Conf. Comput. Vis.})

@String(CVPRW= {IEEE Conf. Comput. Vis. Pattern Recog. Worksh.})

@String(CVPR  = {CVPR})

@String(ICCV  = {ICCV})

@String(CVPRW= {CVPRW})

@String{Computing = "Computing" }

@String{Computer = "{IEEE} Computer" }

@String{Springer = "Springer-Verlag" }

@misc{GoogleEarth,
    author = "Google",
    title = {{Google Earth}},
    note = {{https://earth.google.com}},
    addendum = "(accessed: 16.07.2025)", 
    year = {2025}
}

@misc{EarthEngineAPI,
    author = "Guillaume Attard",
    title = {{An Intro to the Earth Engine Python API}},
    note = {{https://developers.google.com/earth-engine/tutorials/community/intro-to-python-api}},
    addendum = "(accessed: 23.01.2025)", 
    year = {2025}
}

@article{PoissonMesh,
    author = "Michael Kazhdan and Hugues Hoppe",
    title = {{Screened Poisson Surface Reconstruction}},
    journal = {{ACM Transactions on Graphics}},
    volume = 32,
    doi = {https://doi.org/10.1145/2487228.2487237},
    year = 2013
}

@misc{adaptors_T2I,
    title={T2I-Adapter: Learning Adapters to Dig out More Controllable Ability for Text-to-Image Diffusion Models}, 
      author={Chong Mou and Xintao Wang and Liangbin Xie and Yanze Wu and Jian Zhang and Zhongang Qi and Ying Shan and Xiaohu Qie},
      year={2023},
      eprint={2302.08453},
      archivePrefix={arXiv},
      primaryClass={cs.CV},
      url={https://arxiv.org/abs/2302.08453}, 
}

@INPROCEEDINGS{ControlNet,
  author={Zhang, Lvmin and Rao, Anyi and Agrawala, Maneesh},
  booktitle={2023 IEEE/CVF International Conference on Computer Vision (ICCV)}, 
  title={Adding Conditional Control to Text-to-Image Diffusion Models}, 
  year={2023},
  pages={3813-3824},
  doi={10.1109/ICCV51070.2023.00355}
}

@misc{controlingDiffusion_textPromptOneWord,
      title={An Image is Worth One Word: Personalizing Text-to-Image Generation using Textual Inversion}, 
      author={Rinon Gal and Yuval Alaluf and Yuval Atzmon and Or Patashnik and Amit H. Bermano and Gal Chechik and Daniel Cohen-Or},
      year={2022},
      eprint={2208.01618},
      archivePrefix={arXiv},
      primaryClass={cs.CV},
      url={https://arxiv.org/abs/2208.01618}, 
}

@misc{controlingDiffusion_mask,
      title={SDEdit: Guided Image Synthesis and Editing with Stochastic Differential Equations}, 
      author={Chenlin Meng and Yutong He and Yang Song and Jiaming Song and Jiajun Wu and Jun-Yan Zhu and Stefano Ermon},
      year={2022},
      eprint={2108.01073},
      archivePrefix={arXiv},
      primaryClass={cs.CV},
      url={https://arxiv.org/abs/2108.01073}, 
}

@misc{adaptors_ip,
      title={IP-Adapter: Text Compatible Image Prompt Adapter for Text-to-Image Diffusion Models}, 
      author={Hu Ye and Jun Zhang and Sibo Liu and Xiao Han and Wei Yang},
      year={2023},
      eprint={2308.06721},
      archivePrefix={arXiv},
      primaryClass={cs.CV},
      url={https://arxiv.org/abs/2308.06721}, 
}

@misc{3DGS,
      title={3D Gaussian Splatting for Real-Time Radiance Field Rendering}, 
      author={Bernhard Kerbl and Georgios Kopanas and Thomas Leimkühler and George Drettakis},
      year={2023},
      eprint={2308.04079},
      archivePrefix={arXiv},
      primaryClass={cs.GR},
      url={https://arxiv.org/abs/2308.04079}, 
}

@misc{automatic_3DReconstruction_satellite,
  author={Facciolo, Gabriele and De Franchis, Carlo and Meinhardt-Llopis, Enric},
  booktitle={2017 IEEE Conference on Computer Vision and Pattern Recognition Workshops (CVPRW)}, 
  title={Automatic 3D Reconstruction from Multi-date Satellite Images}, 
  year={2017},
  pages={1542-1551},
  doi={10.1109/CVPRW.2017.198}
}

@mastersthesis{satellite_3DReconstruction,
    school = {Linköping University, Department of Electrical Engineering, Computer Vision},
    author = {Tim Yngesjö},
    title = {3D Reconstruction from Satellite Imagery Using Deep Learning},
    year = {2021}
}

@misc{geo_localization_alps,
    author = {Saurer, Olivier and Baatz, Georges and Köser, Kevin and Ladický, L’ubor and Pollefeys, Marc},
    year = {2015},
    month = {06},
    title = {Image Based Geo-localization in the Alps},
    volume = {116},
    journal = {International Journal of Computer Vision},
    doi = {10.1007/s11263-015-0830-0}
}

@misc{zeng2023paint3dpaint3dlightingless,
      title={Paint3D: Paint Anything 3D with Lighting-Less Texture Diffusion Models}, 
      author={Xianfang Zeng and Xin Chen and Zhongqi Qi and Wen Liu and Zibo Zhao and Zhibin Wang and Bin Fu and Yong Liu and Gang Yu},
      year={2023},
      eprint={2312.13913},
      archivePrefix={arXiv},
      primaryClass={cs.CV},
      url={https://arxiv.org/abs/2312.13913}, 
}

@misc{xatlas_library,
    author = {Jonathan Young},
    title = {xatlas: Mesh parameterization / UV unwrapping library},
    url = {https://github.com/jpcy/xatlas},
    year = {2022}
}

@article{geo_pose_cityScape,
    author = {D. Chen and G. Baatz and Köser and S. Tsai and R. Vedantham and T. Pylvanainen and K. Roimela and X. Chen and J. Bach andM. Pollefeys and B. Girod and R. Grzeszczuk} ,
    title = {City-scale landmark identification on mobile devices},
    journal = {InProceedings of Computer Vision and Pattern Recognition (CVPR)},
    year = {2011}
}

@misc{kulhanek2024wildgaussians3dgaussiansplatting,
      title={WildGaussians: 3D Gaussian Splatting in the Wild}, 
      author={Jonas Kulhanek and Songyou Peng and Zuzana Kukelova and Marc Pollefeys and Torsten Sattler},
      year={2024},
      eprint={2407.08447},
      archivePrefix={arXiv},
      primaryClass={cs.CV},
      url={https://arxiv.org/abs/2407.08447}, 
}

@article{ssmi_score,
    author = {Zhou Wang and Eero P Simoncelli and Alan C Bovik},
    title = {Multiscale structural similarity for image quality assessment},
    journal = {The Thrity-Seventh Asilomar Conference on Signals, Systems \& Computers},
    year = {2003},
    pages = {1398–1402},
    volume = {2}
}

@article{LPIPS_score,
    author = {Richard Zhang and Phillip Isola and Alexei A Efros and Eli Shechtman and Oliver Wang},
    title = {The unreasonable effectiveness of deep features as a perceptual metric},
    journal = {CVPR},
    year = {2018}
}

@misc{oquab2024dinov2learningrobustvisual,
      title={DINOv2: Learning Robust Visual Features without Supervision}, 
      author={Maxime Oquab and Timothée Darcet and Théo Moutakanni and Huy Vo and Marc Szafraniec and Vasil Khalidov and Pierre Fernandez and Daniel Haziza and Francisco Massa and Alaaeldin El-Nouby and Mahmoud Assran and Nicolas Ballas and Wojciech Galuba and Russell Howes and Po-Yao Huang and Shang-Wen Li and Ishan Misra and Michael Rabbat and Vasu Sharma and Gabriel Synnaeve and Hu Xu and Hervé Jegou and Julien Mairal and Patrick Labatut and Armand Joulin and Piotr Bojanowski},
      year={2024},
      eprint={2304.07193},
      archivePrefix={arXiv},
      primaryClass={cs.CV},
      url={https://arxiv.org/abs/2304.07193}, 
}

@misc{nerf_novel_view,
      title={NeRF: Representing Scenes as Neural Radiance Fields for View Synthesis}, 
      author={Ben Mildenhall and Pratul P. Srinivasan and Matthew Tancik and Jonathan T. Barron and Ravi Ramamoorthi and Ren Ng},
      year={2020},
      eprint={2003.08934},
      archivePrefix={arXiv},
      primaryClass={cs.CV},
      url={https://arxiv.org/abs/2003.08934}, 
}

@misc{novel_view_mip_nerf,
      title={Mip-NeRF 360: Unbounded Anti-Aliased Neural Radiance Fields}, 
      author={Jonathan T. Barron and Ben Mildenhall and Dor Verbin and Pratul P. Srinivasan and Peter Hedman},
      year={2022},
      eprint={2111.12077},
      archivePrefix={arXiv},
      primaryClass={cs.CV},
      url={https://arxiv.org/abs/2111.12077}, 
}

@misc{novel_view_mipSplatting,
      title={Mip-Splatting: Alias-free 3D Gaussian Splatting}, 
      author={Zehao Yu and Anpei Chen and Binbin Huang and Torsten Sattler and Andreas Geiger},
      year={2023},
      eprint={2311.16493},
      archivePrefix={arXiv},
      primaryClass={cs.CV},
      url={https://arxiv.org/abs/2311.16493}, 
}

@misc{nerf_on_the_go,
      title={NeRF On-the-go: Exploiting Uncertainty for Distractor-free NeRFs in the Wild}, 
      author={Weining Ren and Zihan Zhu and Boyang Sun and Jiaqi Chen and Marc Pollefeys and Songyou Peng},
      year={2024},
      eprint={2405.18715},
      archivePrefix={arXiv},
      primaryClass={cs.CV},
      url={https://arxiv.org/abs/2405.18715}, 
}

@misc{texture_traditional1,
      title={Adversarial Texture Optimization from RGB-D Scans}, 
      author={Jingwei Huang and Justus Thies and Angela Dai and Abhijit Kundu and Chiyu Max Jiang and Leonidas Guibas and Matthias Nießner and Thomas Funkhouser},
      year={2020},
      eprint={2003.08400},
      archivePrefix={arXiv},
      primaryClass={cs.CV},
      url={https://arxiv.org/abs/2003.08400}, 
}

@article{texture_traditional2,
    author  = {Johannes Kopf and Chi-Wing Fu and Daniel Cohen-Or and 
               Oliver Deussen and Dani Lischinski and Tien-Tsin Wong},
    title   = {Solid Texture Synthesis from 2D Exemplars},
    journal = {ACM Transactions on Graphics (Proceedings of SIGGRAPH 2007)},
    year    = {2007},
    volume  = {26},
    number  = {3},
    pages   = {2:1--2:9},
}

@inproceedings{texture_traditional3,
    author = {Turk, Greg},
    title = {Texture synthesis on surfaces},
    year = {2001},
    isbn = {158113374X},
    publisher = {Association for Computing Machinery},
    address = {New York, NY, USA},
    url = {https://doi.org/10.1145/383259.383297},
    doi = {10.1145/383259.383297},
    booktitle = {Proceedings of the 28th Annual Conference on Computer Graphics and Interactive Techniques},
    pages = {347–354},
    numpages = {8},
    series = {SIGGRAPH '01}
}

@misc{texture_learning1,
      title={IT3D: Improved Text-to-3D Generation with Explicit View Synthesis}, 
      author={Yiwen Chen and Chi Zhang and Xiaofeng Yang and Zhongang Cai and Gang Yu and Lei Yang and Guosheng Lin},
      year={2023},
      eprint={2308.11473},
      archivePrefix={arXiv},
      primaryClass={cs.CV},
      url={https://arxiv.org/abs/2308.11473}, 
}

@misc{texture_learning2,
      title={HoloFusion: Towards Photo-realistic 3D Generative Modeling}, 
      author={Animesh Karnewar and Niloy J. Mitra and Andrea Vedaldi and David Novotny},
      year={2023},
      eprint={2308.14244},
      archivePrefix={arXiv},
      primaryClass={cs.CV},
      url={https://arxiv.org/abs/2308.14244}, 
}

@misc{texture_learning3,
      title={Magic123: One Image to High-Quality 3D Object Generation Using Both 2D and 3D Diffusion Priors}, 
      author={Guocheng Qian and Jinjie Mai and Abdullah Hamdi and Jian Ren and Aliaksandr Siarohin and Bing Li and Hsin-Ying Lee and Ivan Skorokhodov and Peter Wonka and Sergey Tulyakov and Bernard Ghanem},
      year={2023},
      eprint={2306.17843},
      archivePrefix={arXiv},
      primaryClass={cs.CV},
      url={https://arxiv.org/abs/2306.17843}, 
}

@misc{texture_learning4,
      title={Make-It-3D: High-Fidelity 3D Creation from A Single Image with Diffusion Prior}, 
      author={Junshu Tang and Tengfei Wang and Bo Zhang and Ting Zhang and Ran Yi and Lizhuang Ma and Dong Chen},
      year={2023},
      eprint={2303.14184},
      archivePrefix={arXiv},
      primaryClass={cs.CV},
      url={https://arxiv.org/abs/2303.14184}, 
}

@misc{texture_3d,
      title={TEXTure: Text-Guided Texturing of 3D Shapes}, 
      author={Elad Richardson and Gal Metzer and Yuval Alaluf and Raja Giryes and Daniel Cohen-Or},
      year={2023},
      eprint={2302.01721},
      archivePrefix={arXiv},
      primaryClass={cs.CV},
      url={https://arxiv.org/abs/2302.01721}, 
}

@misc{textFusion,
      title={TexFusion: Synthesizing 3D Textures with Text-Guided Image Diffusion Models}, 
      author={Tianshi Cao and Karsten Kreis and Sanja Fidler and Nicholas Sharp and Kangxue Yin},
      year={2023},
      eprint={2310.13772},
      archivePrefix={arXiv},
      primaryClass={cs.CV},
      url={https://arxiv.org/abs/2310.13772}, 
}

@misc{text2text,
      title={Text2Tex: Text-driven Texture Synthesis via Diffusion Models}, 
      author={Dave Zhenyu Chen and Yawar Siddiqui and Hsin-Ying Lee and Sergey Tulyakov and Matthias Nießner},
      year={2023},
      eprint={2303.11396},
      archivePrefix={arXiv},
      primaryClass={cs.CV},
      url={https://arxiv.org/abs/2303.11396}, 
}

@inproceedings{lazova2023control,
  title={Control-nerf: Editable feature volumes for scene rendering and manipulation},
  author={Lazova, Verica and Guzov, Vladimir and Olszewski, Kyle and Tulyakov, Sergey and Pons-Moll, Gerard},
  booktitle={Proceedings of the IEEE/CVF Winter Conference on Applications of Computer Vision},
  pages={4340--4350},
  year={2023}
}

@inproceedings{kania2022conerf,
  title={Conerf: Controllable neural radiance fields},
  author={Kania, Kacper and Yi, Kwang Moo and Kowalski, Marek and Trzci{\'n}ski, Tomasz and Tagliasacchi, Andrea},
  booktitle={Proceedings of the IEEE/CVF Conference on Computer Vision and Pattern Recognition},
  pages={18623--18632},
  year={2022}
}

@inproceedings{wang2022clip,
  title={Clip-nerf: Text-and-image driven manipulation of neural radiance fields},
  author={Wang, Can and Chai, Menglei and He, Mingming and Chen, Dongdong and Liao, Jing},
  booktitle={Proceedings of the IEEE/CVF Conference on Computer Vision and Pattern Recognition},
  pages={3835--3844},
  year={2022}
}

@misc{zhang2024coarfcontrollable3dartistic,
      title={CoARF: Controllable 3D Artistic Style Transfer for Radiance Fields}, 
      author={Deheng Zhang and Clara Fernandez-Labrador and Christopher Schroers},
      year={2024},
      eprint={2404.14967},
      archivePrefix={arXiv},
      primaryClass={cs.CV},
      url={https://arxiv.org/abs/2404.14967}, 
}

@inproceedings{zhao2024illuminerf,
    author    = {Xiaoming Zhao and Pratul P. Srinivasan and Dor Verbin and Keunhong Park and Ricardo Martin Brualla and Philipp Henzler},
    title     = {{IllumiNeRF: 3D Relighting Without Inverse Rendering}},
    booktitle = {NeurIPS},
    year      = {2024},
}

@inproceedings{jin2024neural_gaffer,
  title     = {Neural Gaffer: Relighting Any Object via Diffusion},
  author    = {Haian Jin and Yuan Li and Fujun Luan and Yuanbo Xiangli and Sai Bi and Kai Zhang and Zexiang Xu and Jin Sun and Noah Snavely},
  booktitle = {Advances in Neural Information Processing Systems},
  year      = {2024},
}

@inproceedings{munkberg2022extracting,
  title={Extracting triangular 3d models, materials, and lighting from images},
  author={Munkberg, Jacob and Hasselgren, Jon and Shen, Tianchang and Gao, Jun and Chen, Wenzheng and Evans, Alex and M{\"u}ller, Thomas and Fidler, Sanja},
  booktitle=CVPR,
  year={2022}
}

@inproceedings{boss2021nerd,
  title         = {NeRD: Neural Reflectance Decomposition from Image Collections},
  author        = {Boss, Mark and Braun, Raphael and Jampani, Varun and Barron, Jonathan T. and Liu, Ce and Lensch, Hendrik P.A.},
  booktitle     = ICCV,
  year          = {2021},
}

@inproceedings{jin2023tensoir,
  title={Tensoir: Tensorial inverse rendering},
  author={Jin, Haian and Liu, Isabella and Xu, Peijia and Zhang, Xiaoshuai and Han, Songfang and Bi, Sai and Zhou, Xiaowei and Xu, Zexiang and Su, Hao},
  booktitle=CVPR,
  year={2023}
}

@inproceedings{liang2024gsir,
  title     = {GS-IR: 3D Gaussian Splatting for Inverse Rendering}, 
  author    = {Zhihao Liang and Qi Zhang and Ying Feng and Ying Shan and Kui Jia},
  year      = {2024},
  booktitle = CVPR
}

@inproceedings{zhang2025rise,
    title={RISE-SDF: A Relightable Information-Shared Signed Distance Field for Glossy Object Inverse Rendering},
    author={Zhang, Deheng and Wang, Jingyu and Wang, Shaofei and Mihajlovic, Marko and Prokudin, Sergey and Lensch, Hendrik and Tang, Siyu},
    booktitle={International Conference on 3D Vision (3DV)},
    year={2025}
}

@misc{wu2025gsssr,
  title={3D Gaussian Inverse Rendering with Approximated Global Illumination},
  author={Wu, Zirui and Chen, Jianteng and Li, Laijian and Wu, Shaoteng and Zhu, Zhikai and Xu, Kang and Oswald, Martin R. and Song, Jie},
  journal={arXiv preprint},
  url={https://arxiv.org/abs/2504.01358},
  year={2025}
}

@misc{liang2024gusirgaussiansplattingunified,
      title={GUS-IR: Gaussian Splatting with Unified Shading for Inverse Rendering}, 
      author={Zhihao Liang and Hongdong Li and Kui Jia and Kailing Guo and Qi Zhang},
      year={2024},
      eprint={2411.07478},
      archivePrefix={arXiv},
      primaryClass={cs.CV},
      url={https://arxiv.org/abs/2411.07478}, 
}

@inproceedings{zhu2023i2,
    title={I2-SDF: Intrinsic Indoor Scene Reconstruction and Editing via Raytracing in Neural SDFs},
    author={Zhu, Jingsen and Huo, Yuchi and Ye, Qi and Luan, Fujun and Li, Jifan and Xi, Dianbing and Wang, Lisha and Tang, Rui and Hua, Wei and Bao, Hujun and others},
    booktitle={Proceedings of the IEEE/CVF Conference on Computer Vision and Pattern Recognition},
    pages={12489--12498},
    year={2023}
}

@misc{yu2023pointbased,
      title={Point-Based Radiance Fields for Controllable Human Motion Synthesis}, 
      author={Haitao Yu and Deheng Zhang and Peiyuan Xie and Tianyi Zhang},
      year={2023},
      eprint={2310.03375},
      archivePrefix={arXiv},
      primaryClass={cs.CV}
}

@inproceedings{martinbrualla2020nerfw,
    author = {Martin-Brualla, Ricardo
                and Radwan, Noha
                and Sajjadi, Mehdi S. M.
                and Barron, Jonathan T.
                and Dosovitskiy, Alexey
                and Duckworth, Daniel},
    title = {{NeRF in the Wild: Neural Radiance Fields for
            Unconstrained Photo Collections}},
    booktitle = {CVPR},
    year={2021}
}

@inproceedings{li2023matrixcity,
  title={Matrixcity: A large-scale city dataset for city-scale neural rendering and beyond},
  author={Li, Yixuan and Jiang, Lihan and Xu, Linning and Xiangli, Yuanbo and Wang, Zhenzhi and Lin, Dahua and Dai, Bo},
  booktitle={Proceedings of the IEEE/CVF International Conference on Computer Vision},
  pages={3205--3215},
  year={2023}
}

@InProceedings{Turki_2022_CVPR,
    author    = {Turki, Haithem and Ramanan, Deva and Satyanarayanan, Mahadev},
    title     = {Mega-NERF: Scalable Construction of Large-Scale NeRFs for Virtual Fly-Throughs},
    booktitle = {Proceedings of the IEEE/CVF Conference on Computer Vision and Pattern Recognition (CVPR)},
    month     = {June},
    year      = {2022},
    pages     = {12922-12931}
}

@article{Geiger2013IJRR,
  author = {Andreas Geiger and Philip Lenz and Christoph Stiller and Raquel Urtasun},
  title = {Vision meets Robotics: The KITTI Dataset},
  journal = {International Journal of Robotics Research (IJRR)},
  year = {2013}
}

@Article{hierarchicalgaussians24,
      author       = {Kerbl, Bernhard and Meuleman, Andreas and Kopanas, Georgios and Wimmer, Michael and Lanvin, Alexandre and Drettakis, George},
      title        = {A Hierarchical 3D Gaussian Representation for Real-Time Rendering of Very Large Datasets},
      journal      = {ACM Transactions on Graphics},
      number       = {4},
      volume       = {43},
      month        = {July},
      year         = {2024},
      url          = {https://repo-sam.inria.fr/fungraph/hierarchical-3d-gaussians/}
}

@inproceedings{liu2025citygaussian,
  title={Citygaussian: Real-time high-quality large-scale scene rendering with gaussians},
  author={Liu, Yang and Luo, Chuanchen and Fan, Lue and Wang, Naiyan and Peng, Junran and Zhang, Zhaoxiang},
  booktitle={European Conference on Computer Vision},
  pages={265--282},
  year={2025},
  organization={Springer}
}

@misc{ma2025cityloc,
      title={CityLoc: 6DoF Pose Distributional Localization for Text Descriptions in Large-Scale Scenes with Gaussian Representation},
      author={Ma, Qi and Yang, Runyi and Ren, Bin and Sebe, Nicu and Konukoglu, Ender and Van Gool, Luc and Paudel, Danda Pani},
      journal={arXiv preprint arXiv:2501.08982},
      year={2025}
    }

@misc{feng2025citygptempoweringurbanspatial,
      title={CityGPT: Empowering Urban Spatial Cognition of Large Language Models}, 
      author={Jie Feng and Tianhui Liu and Yuwei Du and Siqi Guo and Yuming Lin and Yong Li},
      year={2025},
      eprint={2406.13948},
      archivePrefix={arXiv},
      primaryClass={cs.AI},
      url={https://arxiv.org/abs/2406.13948}, 
}

@inproceedings{ho2020denoising,
  title={Denoising diffusion probabilistic models},
  author={Ho, Jonathan and Jain, Ajay and Abbeel, Pieter},
  booktitle={NeurIPS},
  pages={6840--6851},
  year={2020}
}

@inproceedings{sohl2015deep,
  title={Deep unsupervised learning using nonequilibrium thermodynamics},
  author={Sohl-Dickstein, Jascha and Weiss, Eric and Maheswaranathan, Niru and Ganguli, Surya},
  booktitle={ICML},
  pages={2256--2265},
  year={2015}
}

@article{wildscenes2024,
      title={WildScenes: A Benchmark for 2D and 3D Semantic Segmentation in Large-scale Natural Environments}, 
      author={Kavisha Vidanapathirana and Joshua Knights and Stephen Hausler and Mark Cox and Milad Ramezani and Jason Jooste and Ethan Griffiths and Shaheer Mohamed and Sridha Sridharan and Clinton Fookes and Peyman Moghadam},
      journal = {The International Journal of Robotics Research},
      volume = {44},
      number = {4},
      pages = {532-549},
      year = {2025},
      doi = {10.1177/02783649241278369}
}

@inproceedings{namin2015multi,
  title={A multi-modal graphical model for scene analysis},
  author={Namin, Sarah Taghavi and Najafi, Mohammad and Salzmann, Mathieu and Petersson, Lars},
  booktitle={2015 IEEE Winter Conference on Applications of Computer Vision},
  pages={1006--1013},
  year={2015},
  organization={IEEE}
}

@inproceedings{brejcha2020landscapear,
  title={Landscapear: Large scale outdoor augmented reality by matching photographs with terrain models using learned descriptors},
  author={Brejcha, Jan and Luk{\'a}{\v{c}}, Michal and Hold-Geoffroy, Yannick and Wang, Oliver and {\v{C}}ad{\'\i}k, Martin},
  booktitle={European Conference on Computer Vision},
  pages={295--312},
  year={2020},
  organization={Springer}
}

@inproceedings{RUGD2019IROS,
  author = {Wigness, Maggie and Eum, Sungmin and Rogers, John G and Han, David and Kwon, Heesung},
  title = {A RUGD Dataset for Autonomous Navigation and Visual Perception in Unstructured Outdoor Environments},
  booktitle = {International Conference on Intelligent Robots and Systems (IROS)},
  year = {2019}
}

@misc{jiang2020rellis3d,
      title={RELLIS-3D Dataset: Data, Benchmarks and Analysis}, 
      author={Peng Jiang and Philip Osteen and Maggie Wigness and Srikanth Saripalli},
      year={2020},
      eprint={2011.12954},
      archivePrefix={arXiv},
      primaryClass={cs.CV}
}

@article{zheng2024gaussiangrasper,
      title={GaussianGrasper: 3D Language Gaussian Splatting for Open-vocabulary Robotic Grasping},
      author={Zheng, Yuhang and Chen, Xiangyu and Zheng, Yupeng and Gu, Songen and Yang, Runyi and Jin, Bu and Li, Pengfei and Zhong, Chengliang and Wang, Zengmao and Liu, Lina and others},
      journal={arXiv preprint arXiv:2403.09637},
      year={2024}}
}


\end{document}